\documentclass[10pt,journal,compsoc]{IEEEtran}
\usepackage{amsmath,amsfonts}
\usepackage{algorithmic}
\usepackage{algorithm}
\usepackage{array}
\usepackage[caption=false,font=normalsize,labelfont=sf,textfont=sf]{subfig}
\usepackage{textcomp}
\usepackage{stfloats}
\usepackage{url}
\usepackage{verbatim}
\usepackage{graphicx}
\usepackage{longtable}
\usepackage{cite}
\usepackage{hyperref}
\usepackage{multirow}
\usepackage{booktabs}
\usepackage{lineno}
\usepackage{tikz}
\usetikzlibrary{arrows.meta, positioning, shapes.geometric, calc}
\usepackage{pifont} 

\usepackage{booktabs}
\usepackage{caption}
\usepackage[most]{tcolorbox}
\usepackage{amsmath}
\usepackage{algorithm}
\usepackage{algorithmic}
\begin{document}

\title{A Lightweight Multimodal Vision-Language Framework for Early-Stage Anatomical Green Fruit Classification in Commercial Orchards}

\author{%
Ranjan Sapkota$^{1*}$\thanks{$^{1}$ Cornell University, Department of Environmental and Biological Engineering, USA},
William Bu$^{2}$\thanks{$^{2}$  Department of Computer Science, University of Central Florida, USA},
Chen Chen$^{3}$\thanks{$^{3}$Institute of Artificial Intelligence (IAI) \& Department of Computer Science, University of Central Florida, USA},
Yunjun Xu$^{4}$\thanks{$^{4}$UCF Department of Mechanical and Aerospace Engineering, University of Central Florida, USA},
Manoj Karkee$^{1*}$\thanks{Corresponding author:\texttt{mk2684@cornell.edu}}
}

\maketitle
\begin{abstract} 
Accurate identification of early-stage apple fruitlet anatomical structures, calyx, fruitlet body, and peduncle, is essential for robotic thinning, crop-load management, and other precision management operations in orchards. This study presents a lightweight, multimodal vision–language framework that adapts TinyCLIP for fine-grained fruitlet anatomy classification in complex orchard environments. A dataset of 600 high-resolution RGB images collected from Scilate and Scifresh orchards was converted into $224\times224$ image patches and annotated for three anatomical classes. Domain-specific language prompts (``a photo of a \{class\}'') were used to guide multimodal alignment between orchard imagery and horticultural structures. A sliding-window inference strategy (stride = 112) aggregates patch-level predictions into spatial heatmaps, enabling interpretable whole-image localization of fruitlet components relevant for robotic thinning. Patch-level evaluation on an NVIDIA T4 GPU achieved F1-scores of 0.95 for calyx, 0.98 for fruitlet, and 0.85 for peduncle (macro-F1 = 0.93). Deployment-oriented optimization using ONNX and TensorRT enabled efficient inference on NVIDIA Jetson hardware, preserving accuracy under INT8 quantization while supporting $\sim$127–137 MB model size and millisecond-level patch inference. These results demonstrate that lightweight vision–language models can provide interpretable, edge-deployable perception for automated fruitlet analysis and future robotic thinning systems. The source code and implementation details are publicly available in the project’s GitHub repository at: (\href{https://github.com/WilliamBu1/A-Lightweight-Vision-Language-Model-for-Early-Stage-Fruitlet-Classification-in-Apple-Orchards}{Source Link})
\end{abstract}

\begin{IEEEkeywords}
Agricultural Automation, Greenfruit Thinning, Fruitlet Thinning, Patch-based Image Analysis,  Multi-label Greenfruit Classification
\end{IEEEkeywords}

\section{Introduction}
\begin{figure}[h!]
\centering
\includegraphics[width=0.98\linewidth]{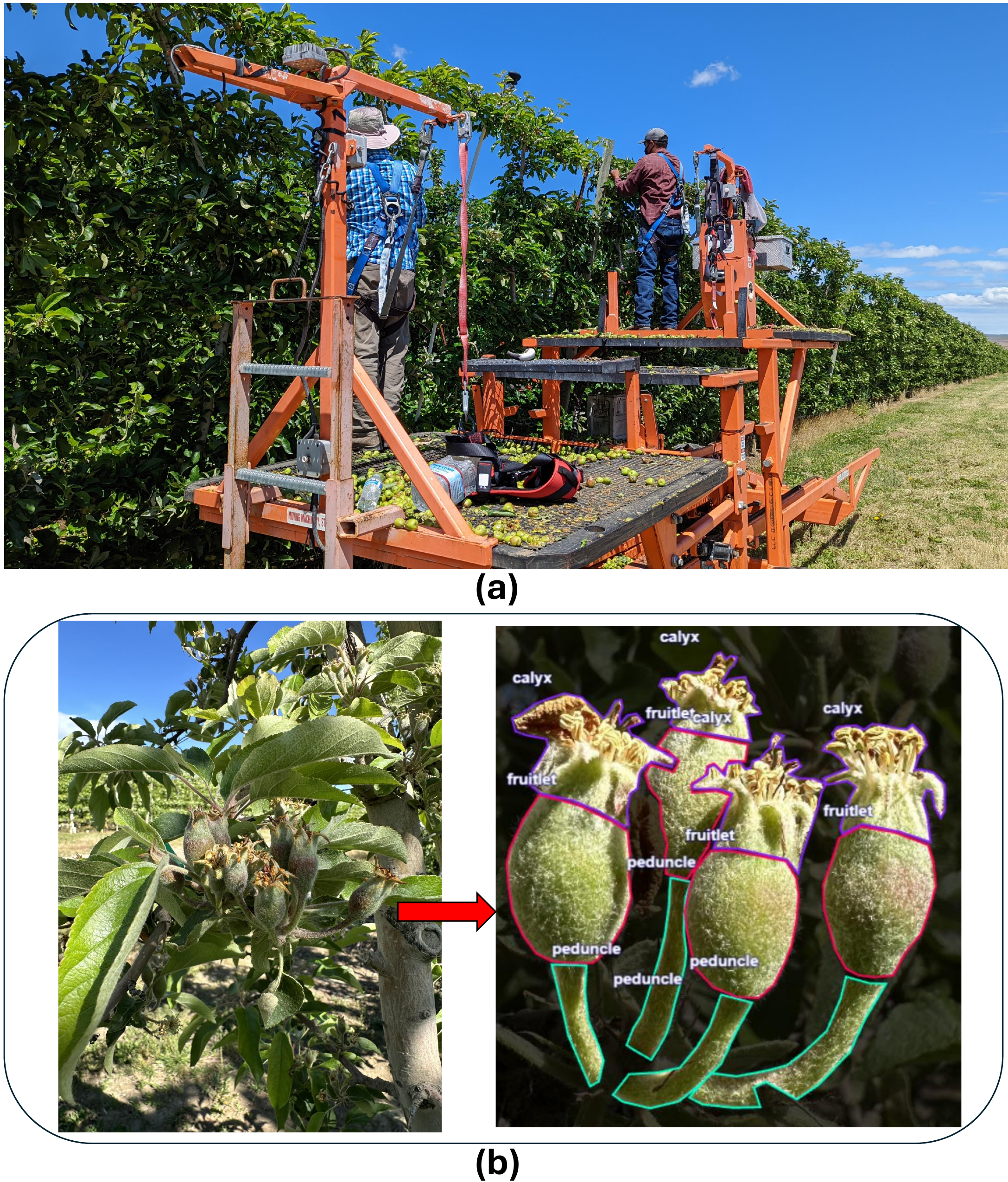}
\caption{(a) Manual thinning of immature green fruitlets in a commercial apple orchard in Washington State. (b) Complex orchard scene with camouflaged early-stage fruitlets and their anatomical components (calyx, fruitlet, peduncle).}
\label{fig:Figure1}
\end{figure}

\textcolor{black}{Early-stage fruitlet thinning is a critical, but labor-intensive operation in commercial apple orchards, which directly influences crop load management, fruit size, and fruit quality \cite{costa2018fruit, greene2012fruit, ouma2012fruit}. Apple trees often overproduce fruitlets \cite{hirst2017advances}, and inadequate thinning leads to overcrowded clusters, reduced fruit size, diminished pack-out quality, and long-term impacts on return bloom \cite{davis2004economics, goffinet1995comparison, sidhu2022crop}. In current practice, thinning is predominantly manual, requiring workers to repeatedly inspect and selectively remove immature fruitlets from dense canopies. As illustrated in Fig.~\ref{fig:Figure1}a, manual thinning is time-consuming and physically demanding, contributing to high labor costs and operational bottlenecks. These challenges are further intensified by global agricultural labor shortages \cite{prause2021digital, sims2017sustainable} and the physical strain associated with repetitive overhead work, which is linked to musculoskeletal disorders and occupational injuries among farm workers \cite{callea2014occupational, fathallah2010musculoskeletal}.}

\textcolor{black}{Automating early-stage thinning requires reliable perception of fruitlet anatomical structures in highly unstructured orchard environments. Early-stage fruitlets are small, green, and often partially occluded by leaves and branches, while their color and texture closely resemble surrounding foliage. Illumination conditions also vary significantly due to shadows, sunflecks, and canopy geometry. Fig.~\ref{fig:Figure1}b illustrates a representative orchard scene where fruitlet anatomical components—calyx, fruitlet, and peduncle—are camouflaged within dense foliage. Within a single fruit cluster, fruitlets may also appear at different developmental stages, further complicating detection and classification. Consequently, perception systems for robotic thinning must accurately localize fine-grained anatomical structures while maintaining computationally efficiency for deployment in field robotics.}

\textcolor{black}{Object detection and localization are fundamental problems in computer vision with applications across robotics, surveillance, medical imaging, autonomous systems, and precision agriculture \cite{zou2023object, hoffmann2020real, choi2019gaussian, zhou2022learning, shi2025advances, mishra2016study}. Early approaches relied on hand-crafted features and classical machine learning models, including Haar-like features, Histogram of Oriented Gradients (HOG), and Support Vector Machine classifiers \cite{dalal2005histograms, chen2013vehicle, sugiarto2017wood, rosyidi2020object}. Although effective in controlled environments, these methods were sensitive to variations in illumination, scale, and occlusion \cite{zhao2019object, zou2019review}, which are common in orchard scenes.}

\textcolor{black}{Deep learning significantly improved visual perception through convolutional neural networks (CNNs), beginning with large-scale image recognition advances demonstrated by AlexNet \cite{krizhevsky2012imagenet}. Region-based detectors such as R-CNN and Faster R-CNN introduced learnable region proposals and end-to-end training \cite{girshick2015fast, ren2016faster}, while single-stage detectors including YOLO and SSD enabled real-time object detection suitable for embedded systems \cite{redmon2016you, liu2016ssd}. Continued development of detection architectures—including CenterNet, EfficientDet, RetinaNet, Cascade R-CNN, and modern YOLO variants—has further refined the trade-off between speed and accuracy for real-time applications \cite{duan2019centernet, tan2020efficientdet, lin2017focal, cai2018cascade, sapkota2025yolo, terven2023comprehensive}. Transformer-based detectors such as DETR and Deformable DETR have further advanced object localization through attention-based modeling, offering improved global-context representation and long-range dependency modeling compared with the predominantly local feature extraction of CNN-based detectors \cite{carion2020end, zhu2020deformable}. Despite these advantages, purely visual models—including both CNN- and transformer-based detectors—can still struggle when target structures exhibit similar appearances or severe occlusion, as frequently encountered in dense orchard environments \cite{jamali2025context, wang2025trifusenet}.}

\textcolor{black}{Vision–language models (VLMs) have recently emerged as a promising approach for addressing such limitations by aligning visual representations with natural language descriptions \cite{shahmohammadi2024language, chen2024spatialvlm}. CLIP demonstrated that large-scale contrastive training on image–text pairs enables strong cross-domain generalization and zero-shot recognition capabilities \cite{radford2021learning}. Subsequent multimodal models such as OpenCLIP, BLIP, and FLAVA further expanded this paradigm by integrating multimodal representation learning and language-guided reasoning \cite{cherti2023reproducible, li2022blip, singh2022flava}. Grounded VLM frameworks including GLIP and Grounding DINO extended multimodal alignment to object-level localization using language prompts \cite{li2022grounded, ren2024grounding, liu2024grounding}. These approaches demonstrate that incorporating semantic information through language can improve perception in visually ambiguous scenes.}

\textcolor{black}{However, most high-capacity VLMs are computationally intensive and difficult to deploy on embedded hardware commonly used in agricultural robotics \cite{wang2025learning, cheng2024spatialrgpt, xu2025exploring, zhang2024vision, luo2023cheap, li2024vision}. This limitation motivates the development of lightweight vision–language models capable of preserving multimodal reasoning while reducing computational requirements. TinyCLIP addresses this challenge by compressing large CLIP models into compact student architectures through distillation techniques while maintaining multimodal alignment \cite{wu2023tinyclip}. Additional lightweight multimodal frameworks, including MobileCLIP, EVA-CLIP, and Q-CLIP, have further demonstrated the feasibility of deploying vision–language models in resource-constrained environments \cite{vasu2024mobileclip, sun2023eva, mi2025q}.}

\textcolor{black}{Such lightweight multimodal perception systems are particularly promising for precision agriculture and orchard robotics. Early-stage fruitlet thinning must be performed within a limited seasonal window and under highly variable environmental conditions. A practical perception system must therefore detect anatomical structures at fine spatial resolution, handle occlusion and canopy clutter, and operate in real time on embedded hardware. Lightweight vision–language models provide a promising compromise by combining semantic understanding with efficient inference.}

\textcolor{black}{In this study, we address the problem of early-stage apple fruitlet anatomical classification in commercial orchards by adapting TinyCLIP for a patch-based multimodal perception framework. Each orchard image is decomposed into overlapping $224\times224$ patches, and TinyCLIP jointly encodes each patch together with class-specific language prompts representing \textit{calyx}, \textit{fruitlet}, and \textit{peduncle}. Patch–prompt similarities in the shared embedding space produce multi-label predictions that indicate the presence of each anatomical component. These predictions are aggregated into class-specific heatmaps that provide interpretable whole-image localization. By combining multimodal semantic grounding with lightweight inference, the proposed framework enables accurate fruitlet anatomy recognition while remaining suitable for deployment in real-world robotic thinning systems.}

\textbf{Key Contributions}

\textcolor{black}{This work makes the following contributions:}

\textcolor{black}{1. We adapt a lightweight Vision–Language Model (TinyCLIP) for fine-grained anatomical classification of early-stage apple fruitlets (calyx, fruitlet, peduncle) in highly occluded orchard environments.}

\textcolor{black}{2. We design a patch-based sliding-window multimodal inference pipeline that converts TinyCLIP predictions into interpretable anatomical heatmaps for whole-image localization.}

\textcolor{black}{3. We demonstrate that the TinyCLIP-based multimodal framework achieves strong classification performance (macro F1 = 0.93) while remaining deployable on edge hardware such as NVIDIA Jetson devices.}

\textcolor{black}{4. We provide a deployment-oriented evaluation comparing FP16 and INT8 TensorRT optimization, showing that lightweight VLMs can achieve real-time inference for robotic orchard perception.}

\textcolor{black}{5. We analyze the effectiveness of multimodal semantic grounding compared with purely visual baselines, demonstrating improved performance for challenging structures such as fruitlet peduncles.}

\section{Methodology}

As depicted in Figure \ref{fig:FigureMethod1}a, We first collect high-res orchard images of Scifresh and Scilate. Experts annotate calyx, fruitlet, and peduncle with COCO boxes in Roboflow. To turn whole images into training examples, we apply a 224×224 sliding window (stride 112) and assign each patch a multi-label vector from overlap with the boxes; background-only areas serve as synthetic negatives. We fine-tune TinyCLIP by aligning patch embeddings with class prompts (“a photo of a {class}”) using a sigmoid head and binary cross-entropy. Training uses AdamW, batch 32, five epochs; no augmentations applied. At inference, we slide and batch patches (up to 64 per step) and aggregate predicted probabilities into class-specific heatmaps for interpretable localization. For edge deployment, the PyTorch model is exported to ONNX and compiled to TensorRT (FP16/INT8). We benchmark latency, memory, and throughput on a T4 GPU and Jetson hardware platforms.

\subsection{Study Site and Data Acquisition}
\begin{figure*}[ht!]
    \centering
    \includegraphics[width=0.78\linewidth]{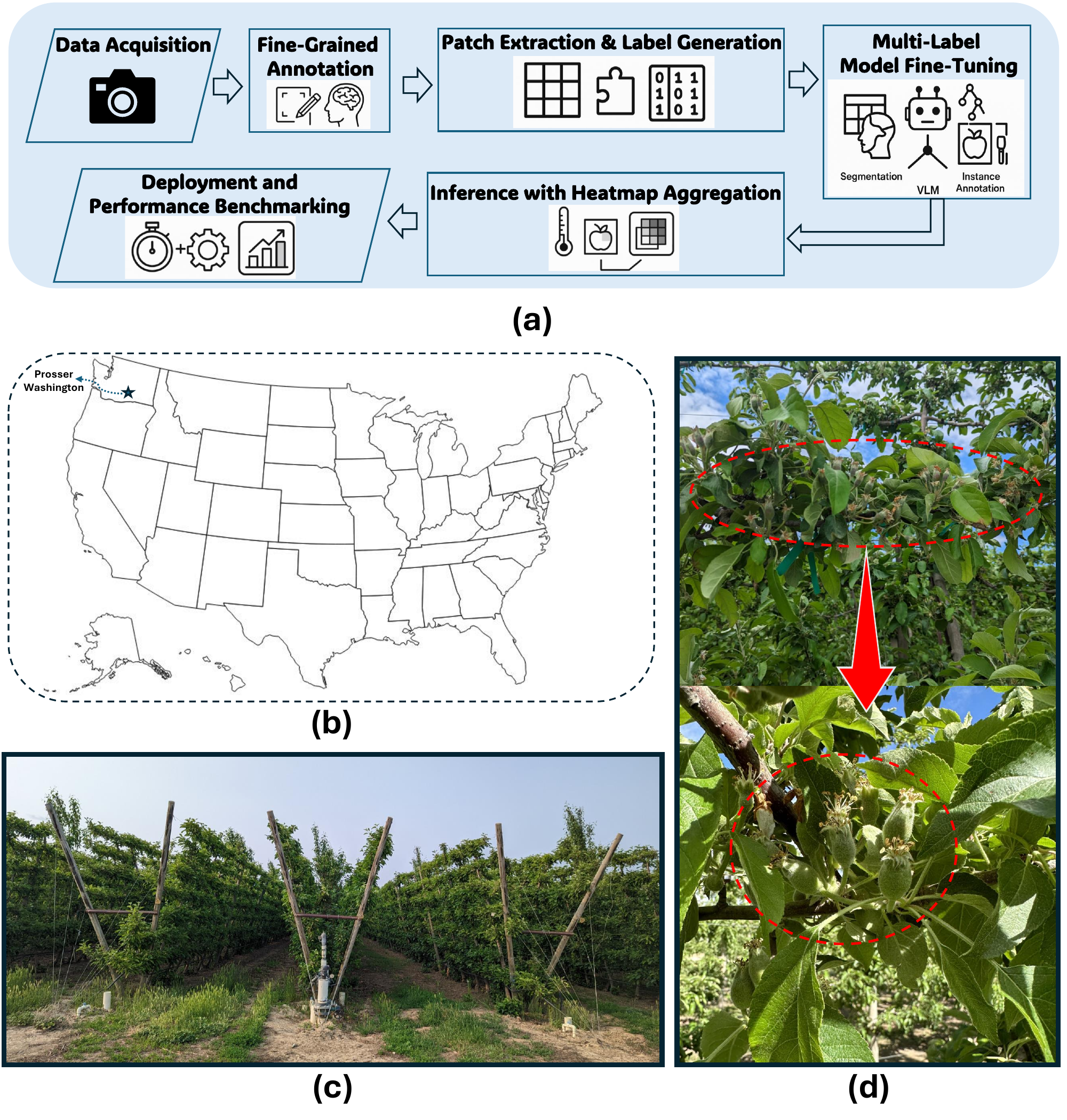}
    \caption{Visual illustration of the orchard setting and imaging conditions. The left panel shows the high-density planting arrangement of Scifresh and Scilate cultivars, while the right panel highlights immature fruitlet clusters captured at varying distances and angles with an iPhone 14 Pro.}
    \label{fig:FigureMethod1}
\end{figure*}

This research was conducted in a commercial apple orchard located in Prosser, Washington State,  USA (example Figure \ref{fig:FigureMethod1}b and \ref{fig:FigureMethod1}c) during the early post-bloom period of June~2024. The orchard comprised two apple cultivars Scifresh and Scilate arranged in high-density rows with approximately 3~ft intra-row spacing and a maintained canopy height of about 10~ft. Photographing immature green fruitlet clusters required capturing diverse perspectives across multiple trees and rows. We collected 600 high-resolution RGB images using an iPhone~14~Pro under natural daylight, varying camera–fruitlet distance (typically within 3~ft) and angle to ensure diversity in scale, orientation, and background context.

\subsection{Dataset Preparation and Annotation}

\subsubsection{Image Acquisition and Class Definitions}
Images were collected from commercial apple orchards during the early fruit development stage. Each image contains three fine-grained anatomical components: \textit{calyx}, \textit{fruitlet}, and \textit{peduncle}. These structures are small, highly occluded, and visually similar to surrounding foliage, necessitating precise annotation and localized learning.

\subsubsection{Annotation Procedure}
All images were annotated using the Roboflow interface. Annotators drew bounding boxes around every instance of the three classes, and annotations were exported in COCO format. A horticulture domain expert cross-verified every annotation for anatomical correctness, bounding-box tightness, and positional accuracy.

\subsubsection{Negative Sample Generation}
Because every orchard image contained at least one target anatomical structure, the dataset lacked image-level negatives. To enable presence–absence learning, we generated negative examples by cropping background regions with no bounding-box overlap. These synthetic negatives were assigned the label vector $[0,0,0]$.

\subsubsection{Dataset Splitting}
The dataset was divided into non-overlapping training (80\%), validation (10\%), and test (10\%) sets. Splits preserved class balance and prevented patch leakage across subsets.


\subsection{Patch Extraction and Multi-Label Generation}

\subsubsection{Sliding-Window Patch Extraction}
To localize anatomical structures, each full-resolution image was partitioned using a sliding window of size $224\times224$ pixels and a stride of 112 pixels (50\% overlap). This produced 900 patches per image, ensuring complete coverage and reducing boundary effects.

\subsubsection{Binary Multi-Label Vector Assignment}

For each extracted patch, we assigned a binary vector 
\[
\mathbf{y} = [y_{\text{calyx}}, y_{\text{fruitlet}}, y_{\text{peduncle}}] \in \{0,1\}^3,
\]
based on overlap between the patch and class-specific bounding boxes. Using a COCO-parsing script, $y_c=1$ if any annotated instance of class $c$ overlapped the patch; otherwise $y_c=0$.

\subsubsection{Negative Patch Enrichment}
To reduce false positives and improve abstention, additional background-only patches were extracted from foliage regions. These patches contained no class instances and were labeled $[0,0,0]$, increasing negative-class diversity.


\subsection{Multi-Label Fine-Tuning of TinyCLIP}

\subsubsection{Model Overview}
\begin{figure}[h!]
\centering
\includegraphics[width=0.98\linewidth]{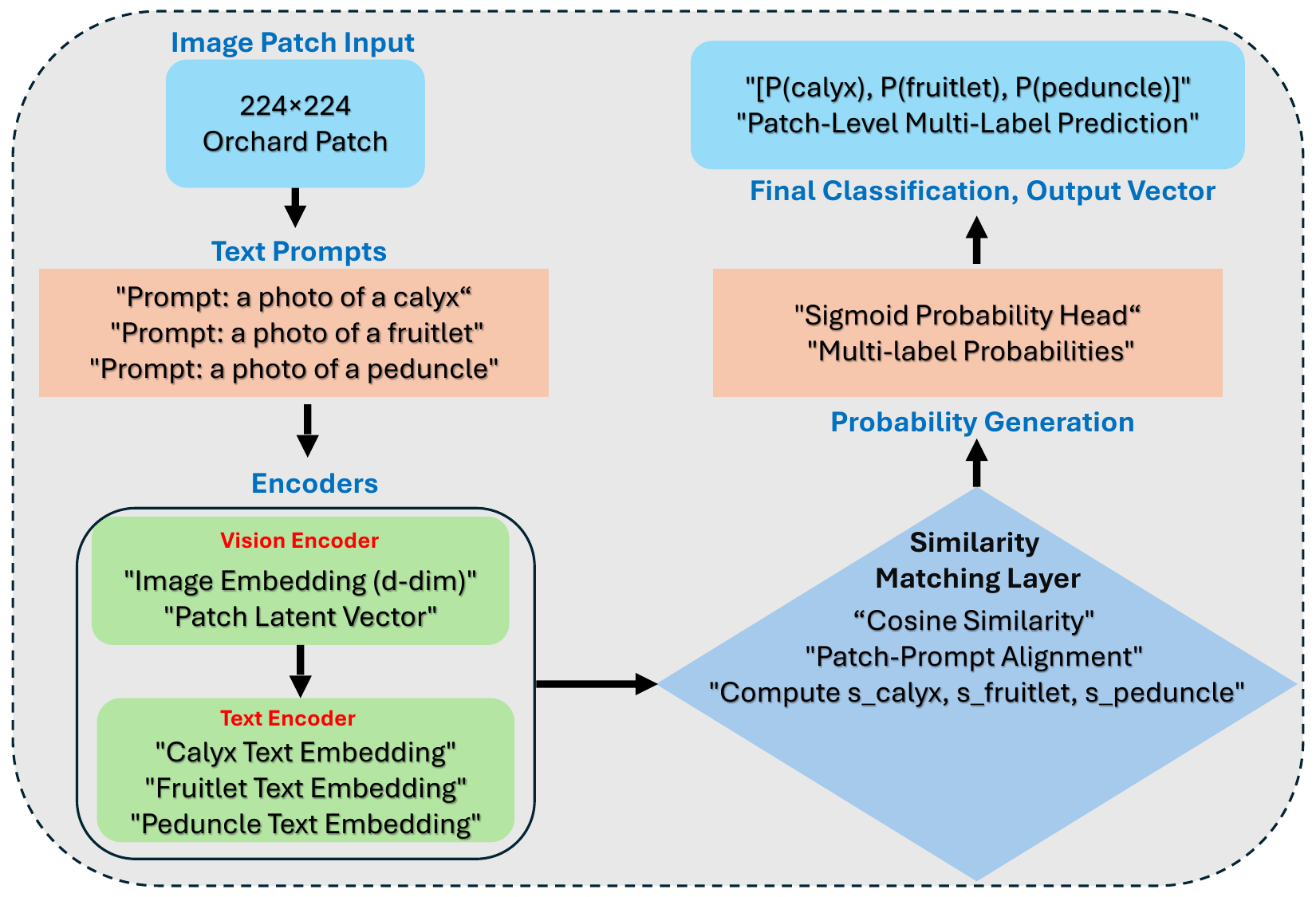}
\caption{Architecture showing the $224\times224$ orchard patch input, TinyCLIP vision encoder, text prompts, text encoder, image and text embeddings, cosine similarity module, sigmoid probability head, and multi-label outputs for calyx, fruitlet, and peduncle classification.}
\label{fig:architecture}
\end{figure}

\textcolor{black}{As shown in Figure~\ref{fig:architecture}, the adapted TinyCLIP framework operates as a dual-encoder multimodal system that embeds both orchard image patches and class-specific natural-language prompts into a shared representation space. Each $224\times224$ image patch extracted from early-season orchard scenes is processed by the TinyCLIP vision encoder to produce a compact visual embedding capturing morphological cues of fruitlet structures under varying illumination and occlusion. In parallel, the TinyCLIP text encoder converts short prompts (e.g., ``a photo of a calyx'', ``a photo of a fruitlet'', and ``a photo of a peduncle'') into corresponding semantic embeddings. } 

\textcolor{black}{The similarity between visual and textual embeddings is computed using cosine similarity, producing class-specific alignment scores. A sigmoid activation then converts these scores into independent multi-label probabilities, allowing the model to predict the presence of multiple anatomical components within each patch. This multimodal formulation enables semantic grounding between visual features and horticultural structures, improving discrimination of visually similar elements such as peduncles and surrounding foliage. The lightweight architecture also supports dense sliding-window evaluation and subsequent heatmap aggregation for whole-image anatomical localization in orchard scenes.}

\begin{figure*}[t]
    \centering
    \includegraphics[width=0.98\linewidth]{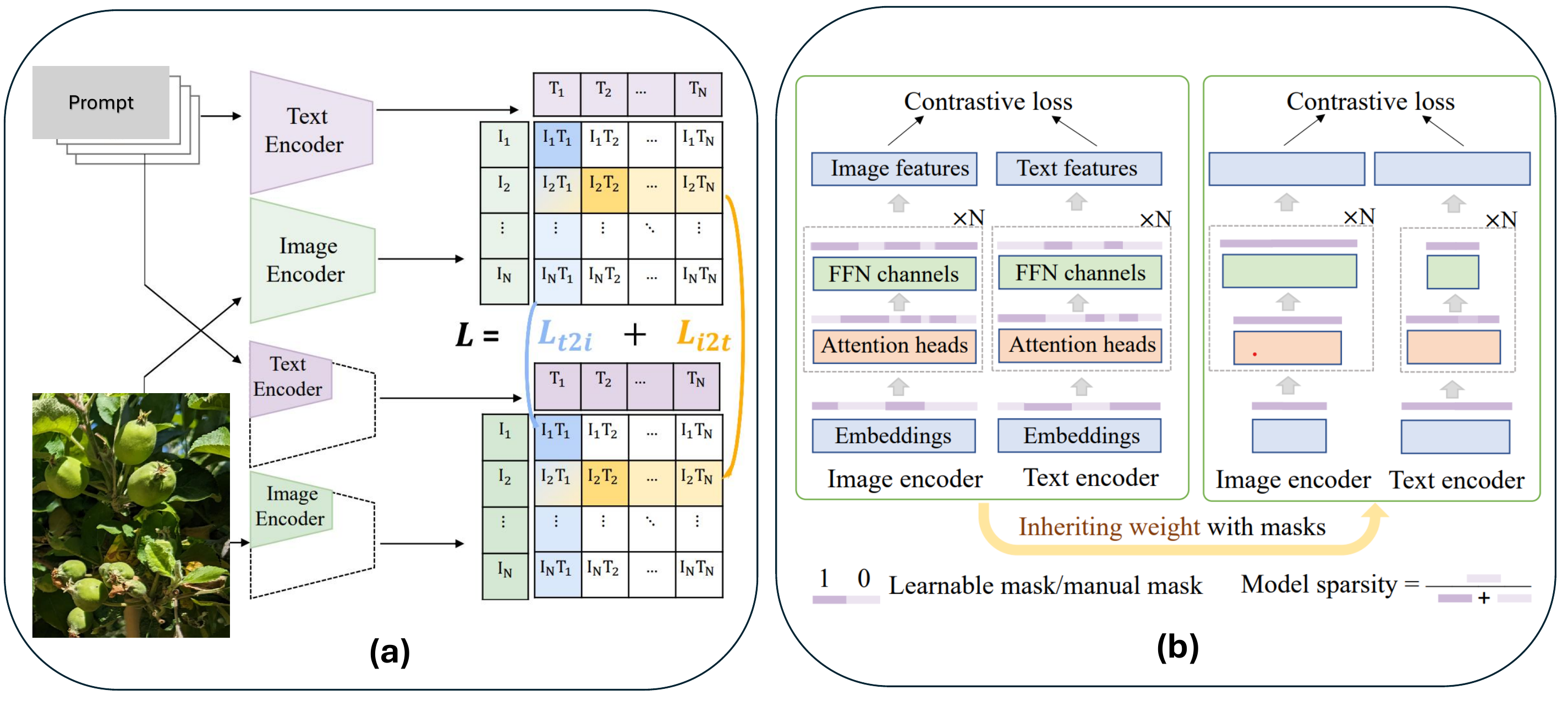}
    \caption{(a) TinyCLIP affinity mimicking: the student distills CLIP’s multimodal geometry by matching image-to-text and text-to-image similarity distributions. (b) Weight inheritance strategy: masks select transferable parameters from a pre-trained CLIP model, enabling compact student initialization while discarding non-essential weights.}
    \label{fig:ARCHITECTURE}
\end{figure*}

\textcolor{black}{TinyCLIP itself is a compact vision–language model distilled from a larger CLIP teacher network and designed for efficient deployment in resource-constrained environments such as embedded robotic systems \cite{wu2023tinyclip}. Figures~\ref{fig:ARCHITECTURE}a and \ref{fig:ARCHITECTURE}b illustrate the original TinyCLIP training concepts, including affinity-based multimodal distillation and weight inheritance from the teacher model. In this study, TinyCLIP is adopted as a pretrained lightweight multimodal backbone and fine-tuned for the specific task of apple fruitlet anatomy classification.}

In our study, we adapt TinyCLIP for multi-label anatomical classification of early-stage apple fruitlet components \textit{calyx}, \textit{fruitlet}, and \textit{peduncle}. Each $224 \times 224$ orchard patch is embedded by the TinyCLIP vision encoder into a compact latent space capturing fine-grained morphological cues. In parallel, the text encoder embeds natural-language prompts of the form ``a photo of a \{class\}'', producing three semantic prototypes corresponding to the anatomical classes. Patch-level classification is then obtained by computing cosine similarity between the patch embedding and each class-specific prompt embedding, effectively grounding visual recognition in a shared linguistic--visual embedding space. This multimodal formulation allows TinyCLIP to differentiate subtle anatomical structures despite occlusions, orchard lighting variability, and small object size, while maintaining computational efficiency suitable for dense sliding-window inference and real-time edge deployment.

\subsubsection{Prompt Engineering}
We used the prompt template:
\[
\texttt{"a photo of a \{class\}"},
\]
substituting \textit{calyx}, \textit{fruitlet}, and \textit{peduncle}. TinyCLIP's text encoder generated three fixed text embeddings, one for each anatomical class.

\subsubsection{Similarity Computation and Probability Mapping}
For a patch embedding $\mathbf{v}$ and class prompt embedding $\mathbf{t}_c$, the cosine similarity output $s_c$ measures alignment strength. A sigmoid activation produced class probabilities:
\[
p_c = \sigma(s_c).
\]

\subsubsection{Training Objective}
We fine-tuned TinyCLIP using Binary Cross-Entropy (BCE) loss:
\[
\mathcal{L}_{\text{BCE}}
=
-\sum_{c=1}^{3}
\left[
y_c \log(p_c) + (1-y_c)\log(1-p_c)
\right].
\]
Optimization used AdamW with a learning rate of $1\times10^{-4}$, weight decay of $1\times10^{-5}$, batch size of 32, and 5 training epochs. No augmentations were applied to evaluate model performance strictly under orchard variability.

\subsubsection{Training Configuration}
The final layers of the vision encoder and the full text encoder were unfrozen. Training was implemented in PyTorch on an NVIDIA T4 GPU. To mitigate class imbalance, training batches were balanced to ensure fair representation of the peduncle class.


\subsection{Sliding-Window Inference and Heatmap Generation}

\subsubsection{Batch Inference}
During inference, the same $224\times224$ sliding-window with stride 112 was applied. Patches were grouped into batches (64 on T4 GPU; up to 8 on Jetson Nano) for parallel processing, significantly reducing inference time.

\subsubsection{Class-Specific Heatmaps}
For each class $c$, a spatial heatmap $H_c(x,y)$ was constructed:
\[
H_c(x,y)
=
\frac{1}{N}
\sum_{i=1}^{N}
p_c^{(i)}\,
\mathbb{I}((x,y)\in W_i),
\]
where $p_c^{(i)}$ is the predicted probability for patch $i$, and $W_i$ is the spatial support of patch $i$. Heatmaps highlight anatomical likelihoods across the image and support decision-making for robotic thinning.


\subsection{Deployment and Inference Efficiency Evaluation}

\subsubsection{Evaluation Metrics}

To rigorously assess the computational performance and deployment feasibility of the proposed TinyCLIP-based fruitlet anatomy classification pipeline, we quantify four key metrics: patch-level latency, full-image inference time, peak GPU memory usage, and throughput. Each metric captures a distinct aspect of the system’s efficiency, and together they characterize its suitability for real-time or near–real-time operation in resource-constrained orchard robotics environments. Below, we provide a detailed scientific description of each metric, accompanied by analytic expressions and explanations tailored to the fruitlet classification context.

\paragraph{1. Patch-Level Latency}
Patch-level latency measures the average time required to process a single $224\times224$ orchard patch through the TinyCLIP model. Here, $t_i$ denote the inference time for patch $i$, and $N$ denote the number of patches in a batch. The average latency is defined as:
\begin{equation}
    \tau_{\text{patch}} = \frac{1}{N} \sum_{i=1}^{N} t_i.
\end{equation}

Here:
\begin{itemize}
    \item $t_i$ represents the forward-pass computation time of the TinyCLIP vision encoder, text encoder lookup, cosine similarity computation, and sigmoid activation for patch $i$.
    \item $N$ is the total number of patches processed in a batch.
\end{itemize}

In the context of fruitlet classification, $\tau_{\text{patch}}$ directly reflects the responsiveness of patch-level decision making. Since each full orchard image is decomposed into dozens of overlapping patches, lower latency ensures that calyx, fruitlet, and peduncle predictions can be generated rapidly enough to guide real-time orchard robots or hand-held devices.

\paragraph{2. Full-Image Inference Time}
Full-image inference time quantifies the total time required to process all sliding-window patches extracted from a single orchard image. Here, $M$ denote the number of patches in the image (dependent on stride and resolution), and let $T$ denote the total processing time. The metric is defined as:
\begin{equation}
    T_{\text{image}} = \sum_{i=1}^{M} t_i.
\end{equation}

Where:
\begin{itemize}
    \item $M$ is the number of overlapping patches extracted via a sliding window (900 patches per image).
    \item $t_i$ is the latency of patch $i$.
\end{itemize}

For fruitlet classification, $T_{\text{image}}$ determines whether the system can keep pace with the movement of a field robot navigating orchard rows. For example, if $T_{\text{image}} < 5$ seconds, a mobile robot can analyze scenes continuously while moving at practical field speeds. Additionally, faster $T_{\text{image}}$ enables high-frequency anatomical heatmap updates for cluster-level decision making.

\paragraph{3. Peak GPU Memory Usage}
Peak GPU memory usage captures the maximum memory required during inference and depends on model size, batch size, activation storage, and TensorRT/ONNX kernel allocations. Formally:
\begin{equation}
    \mathcal{M}_{\text{peak}} = \max_{t} \left( \mathcal{M}_{\text{weights}} + \mathcal{M}_{\text{activations}}(t) + \mathcal{M}_{\text{temporary}}(t) \right),
\end{equation}

Where:
\begin{itemize}
    \item $\mathcal{M}_{\text{weights}}$ is the static memory footprint of TinyCLIP parameters.
    \item $\mathcal{M}_{\text{activations}}(t)$ is the memory required to store intermediate feature maps at time $t$.
    \item $\mathcal{M}_{\text{temporary}}(t)$ includes additional buffers used by TensorRT, ONNX Runtime, or PyTorch kernels during operations.
\end{itemize}

In orchard deployment scenarios, $\mathcal{M}_{\text{peak}}$ dictates compatibility with embedded devices such as NVIDIA Jetson Nano or Xavier NX, which have stringent memory constraints. Lower memory usage is crucial for running sliding-window inference on edge hardware without swapping or thermal throttling. Because fruitlet classification requires evaluating dense grids of patches, memory-efficient inference ensures stable operation even at high patch throughput.

\paragraph{4. Throughput (Images per Second)}
Throughput evaluates how many full orchard images can be processed per second and reflects system-level efficiency. It is computed as the inverse of the full-image inference time:
\begin{equation}
    \Phi = \frac{1}{T_{\text{image}}}.
\end{equation}

Where:
\begin{itemize}
    \item $\Phi$ denotes throughput (images/second).
    \item $T_{\text{image}}$ is defined as above.
\end{itemize}

In the fruitlet classification context, throughput determines whether the TinyCLIP-based system can support real-time orchard monitoring. For autonomous thinning robots, higher $\Phi$ enables continuous perception while traversing orchard rows; for handheld or mobile devices, it ensures responsive user feedback. Throughput also directly affects the generation rate of anatomical heatmaps, which are essential for identifying peduncles (the cutting point) and fruitlets (thinning targets) across full scenes.

Together, these four metrics patch latency ($\tau_{\text{patch}}$), full-image inference time ($T_{\text{image}}$), peak memory usage ($\mathcal{M}_{\text{peak}}$), and throughput ($\Phi$) quantify the computational performance of the TinyCLIP-based fruitlet classification pipeline. Their analytic definitions highlight how each metric influences practical deployment in commercial orchards, where efficient, real-time, and memory-aware perception is critical for enabling next-generation robotic thinning and precision horticultural automation.

\subsubsection{ONNX Export and TensorRT Optimization}
To enable real-time inference on resource-limited embedded platforms such as the Jetson Nano, we converted the fine-tuned TinyCLIP model from PyTorch to the Open Neural Network Exchange (ONNX) format. ONNX serves as an intermediate representation that allows a model trained in one deep-learning framework to be executed efficiently across diverse hardware backends. During export, we enabled dynamic shape support so that the Jetson can flexibly process different numbers of sliding-window patches per image without re-compiling the model. After ONNX export, the model was further optimized using NVIDIA TensorRT, a high-performance inference engine that generates hardware-specific execution plans. A crucial optimization step within TensorRT is \emph{quantization}, which reduces the numerical precision of model weights and activations from 32-bit floating-point to lower-precision formats such as FP16 (16-bit) or INT8 (8-bit). Quantization significantly decreases computation cost, memory usage, and bandwidth requirements while maintaining nearly identical prediction accuracy. In practical terms, FP16 and INT8 TensorRT engines allow the Jetson Nano to run TinyCLIP at faster frame rates and lower energy consumption, making the model suitable for field deployment in orchard robotics. This combination of ONNX portability and TensorRT optimization yields a compact, hardware-accelerated model that meets the latency and memory constraints of embedded systems.

\subsubsection{Batch vs. Sequential Comparison}
We compared two inference strategies sequential and batched to evaluate their computational efficiency on different hardware platforms. In sequential inference, each $224\times224$ patch is processed individually in a separate forward pass through the TinyCLIP model. Although straightforward, this method incurs substantial overhead because the model must repeatedly load intermediate activations and compute similarity scores for each patch. As a result, processing all sliding-window patches from a single orchard image required approximately 20 seconds, which is too slow for real-time robotic applications. Batched inference, by contrast, groups multiple patches into a single tensor and processes them simultaneously in a single forward pass. This approach exploits GPU parallelism and significantly reduces redundant computations. On the NVIDIA T4 GPU, batching up to 64 patches reduced full-image inference time to 3–4 seconds. On the Jetson Nano, where available memory is limited, batch sizes up to 8 patches yielded an inference time of 5–6 seconds per image. Thus, batched inference provides an order-of-magnitude speedup compared to sequential evaluation, enabling responsive anatomical classification and heatmap generation in orchard settings. The optimal batch size depends on hardware memory constraints, making this tuning step essential for efficient deployment.

\subsubsection{Software Environment}
All model training, conversion, and inference experiments were conducted using a stable and reproducible software stack. Python~3.11 served as the programming environment, while PyTorch~2.x provided the primary deep-learning framework for TinyCLIP training and fine-tuning. ONNX Runtime was used to validate exported ONNX models and ensure correctness prior to hardware-level optimization. For Jetson deployment, we utilized NVIDIA TensorRT, which compiles the ONNX graph into optimized FP16 and INT8 engines tailored to the device’s GPU architecture. These engines enable accelerated inference with reduced latency and memory consumption. Profiling tools integrated within PyTorch and TensorRT captured latency measurements, while GPU monitoring utilities monitored memory utilization. All performance metrics reported in this study represent the mean of three independent runs to account for runtime variability. This well-defined software environment ensures that our methodology is reproducible, reliable, and portable across embedded and cloud hardware configurations.

Overall, the deployment methodology combines (1) conversion of the TinyCLIP model to a hardware-agnostic ONNX format, (2) TensorRT quantization and optimization for high-speed inference, (3) batch-optimized patch processing for practical runtime performance, and (4) a robust, well-defined software environment supporting reproducibility. Together, these steps produce a lightweight, efficient, and deployable multimodal perception system capable of performing fine-grained fruitlet anatomy classification in real orchard environments.


\section{Results and Discussion}
\begin{figure*}[h!]
    \centering
    \includegraphics[width= 0.98\linewidth]{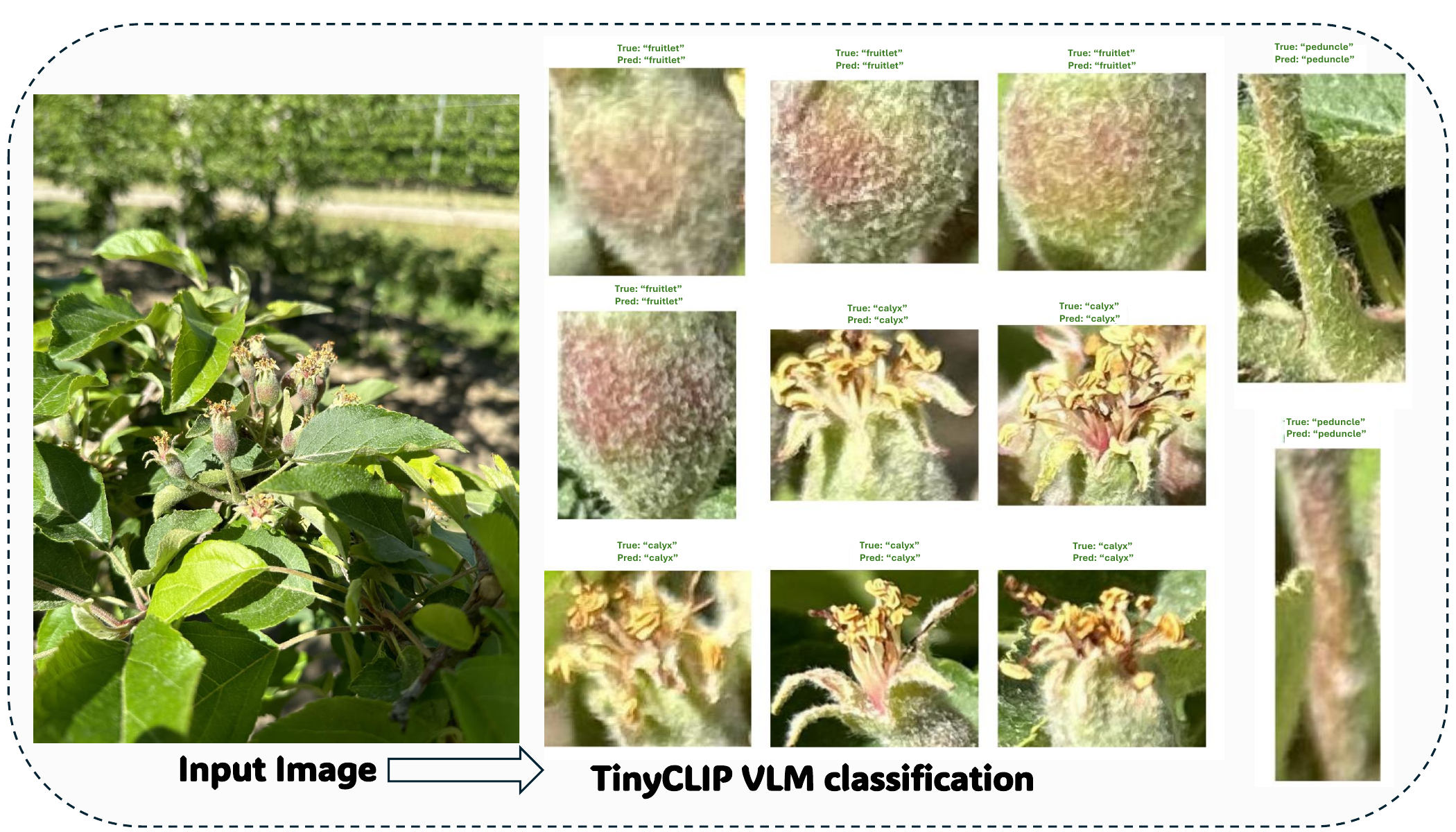}
    \caption{Illustration of TinyCLIP’s multi-modal classification on an early-season orchard image. The left panel shows the original iPhone~14~Pro image captured in a commercial orchard, while the right panel presents representative $224\times224$ patches with corresponding ground-truth (“True”) and predicted (“Pred”) labels for fruitlet, calyx, and peduncle, demonstrating reliable fine-grained anatomical discrimination.}
    \label{fig:RESULTEXAMPLE1}
\end{figure*}

Figure~\ref{fig:RESULTEXAMPLE1} provides a representative example demonstrating TinyCLIP’s ability to correctly classify anatomically distinct fruitlet structures \textit{calyx}, \textit{fruitlet}, and \textit{peduncle} within a highly cluttered and visually challenging orchard scene. The original image on the left of Figure~\ref{fig:RESULTEXAMPLE1} was captured using an iPhone~14~Pro in a commercial apple orchard during the early growing season, a period characterized by dense foliage, substantial occlusion, strong natural illumination variability, and minimal color contrast between fruitlets and the surrounding canopy. These environmental factors typically degrade the performance of conventional purely vision-based detectors and make fine-grained anatomical perception particularly difficult.

On the right side of the figure, representative $224\times224$ patches extracted from the original scene are shown, each annotated with both the ground-truth label (“True”) and the model prediction (“Pred”). These examples illustrate TinyCLIP’s capacity to correctly interpret subtle textural and structural cues: fruitlets are recognized based on their spherical morphology and surface fuzziness, calyx structures are identified through their distinct petal remnants and star-shaped geometry, and peduncles are distinguished by their slender elongated form. The close alignment between “True” and “Pred” labels across these examples underscores the effectiveness of the multimodal architecture in leveraging linguistic priors and visual embeddings for semantic discrimination, even in low-signal, visually congested orchard environments.

This qualitative demonstration establishes the visual intuition underlying the subsequent quantitative analyses, highlighting TinyCLIP’s strong interpretability and fine-grained discriminative capacity prior to the detailed evaluation of classification metrics, localization performance, and deployment efficiency presented in later subsections.

\subsection{Patch-Level Multi-Label Classification Performance}

Patch-level multi-label classification constitutes the core evaluation of the proposed TinyCLIP–based fruitlet anatomy recognition framework. In this experiment, each $224 \times 224$ sliding-window patch is independently evaluated for the presence or absence of calyx, fruitlet, peduncle, and background (negative class). The reported results represent TensorRT-accelerated inference on two optimized TinyCLIP engines: an FP16 engine and an INT8 quantized engine. Each engine was evaluated on 339 patches, with per-class precision, recall, F1-score, support counts, and overall accuracy reported below.

The FP16 engine achieved strong performance across all anatomical classes (Table~\ref{tab:fp16_classification}). For the primary anatomical targets (calyx, fruitlet, peduncle), the FP16 model maintained high precision (0.964–0.988) and strong recall for calyx and fruitlet. Peduncle recall was moderately lower (0.750), indicating occasional missed detections, likely due to its fine, elongated morphology and variable visual presentation under occlusions. The overall F1-score remained high (0.851–0.982), demonstrating the ability of TinyCLIP to adapt to subtle anatomical cues after fine-tuning. The macro-averaged F1-score reached 0.9115, with an overall accuracy of 91.15\%, highlighting the strength of multimodal representations for fine-grained classification tasks.

\begin{table}[h]
\centering
\caption{TensorRT FP16 Classification Metrics (339 patches)}
\label{tab:fp16_classification}
\begin{tabular}{lcccc}
\toprule
\textbf{Class} & \textbf{Precision} & \textbf{Recall} & \textbf{F1} & \textbf{Support} \\
\midrule
Calyx      & 0.9639 & 0.9412 & 0.9524 & 85 \\
Fruitlet   & 0.9881 & 0.9765 & 0.9822 & 85 \\
Peduncle   & 0.9844 & 0.7500 & 0.8514 & 84 \\
Negative   & 0.7685 & 0.9765 & 0.8601 & 85 \\
\midrule
Accuracy   &        &        & 0.9115 & 339 \\
Macro Avg  & 0.9262 & 0.9110 & 0.9115 & 339 \\
Weighted Avg & 0.9260 & 0.9115 & 0.9117 & 339 \\
\bottomrule
\end{tabular}
\end{table}

Quantized INT8 inference exhibited slightly reduced performance (Table~\ref{tab:int8_classification}) but remained competitive, with an accuracy of 88.79\% and macro F1-score of 0.8881. Precision stayed consistently high, demonstrating that INT8 quantization primarily impacts recall rather than false positive rates. This trade-off is typical for highly compressed models but acceptable for presence/absence classification where precision is more important for avoiding erroneous thinning decisions. Notably, peduncle recall decreased to 0.678, further confirming its sensitivity to spatial resolution loss introduced by quantization.

\begin{table}[h]
\centering
\caption{TensorRT INT8 Classification Metrics (339 patches)}
\label{tab:int8_classification}
\begin{tabular}{lcccc}
\toprule
\textbf{Class} & \textbf{Precision} & \textbf{Recall} & \textbf{F1} & \textbf{Support} \\
\midrule
Calyx      & 0.9750 & 0.9176 & 0.9455 & 85 \\
Fruitlet   & 0.9765 & 0.9765 & 0.9765 & 85 \\
Peduncle   & 1.0000 & 0.6786 & 0.8085 & 84 \\
Negative   & 0.7094 & 0.9765 & 0.8218 & 85 \\
\midrule
Accuracy   &        &        & 0.8879 & 339 \\
Macro Avg  & 0.9152 & 0.8873 & 0.8881 & 339 \\
Weighted Avg & 0.9150 & 0.8879 & 0.8883 & 339 \\
\bottomrule
\end{tabular}
\end{table}

Overall, the patch-level analysis confirms that TinyCLIP effectively distinguishes between anatomically similar structures in orchard settings. The use of text prompts significantly improves multi-label classification performance relative to baseline visual models (as shown later in the ablation study), and the model remains robust even under aggressive INT8 quantization. The moderately reduced peduncle recall highlights the importance of integrating contextual spatial information in future iterations, potentially through sequence modeling or region-based VLM extensions.

\begin{figure}[h!]
\centering
\includegraphics[width=0.98\linewidth]{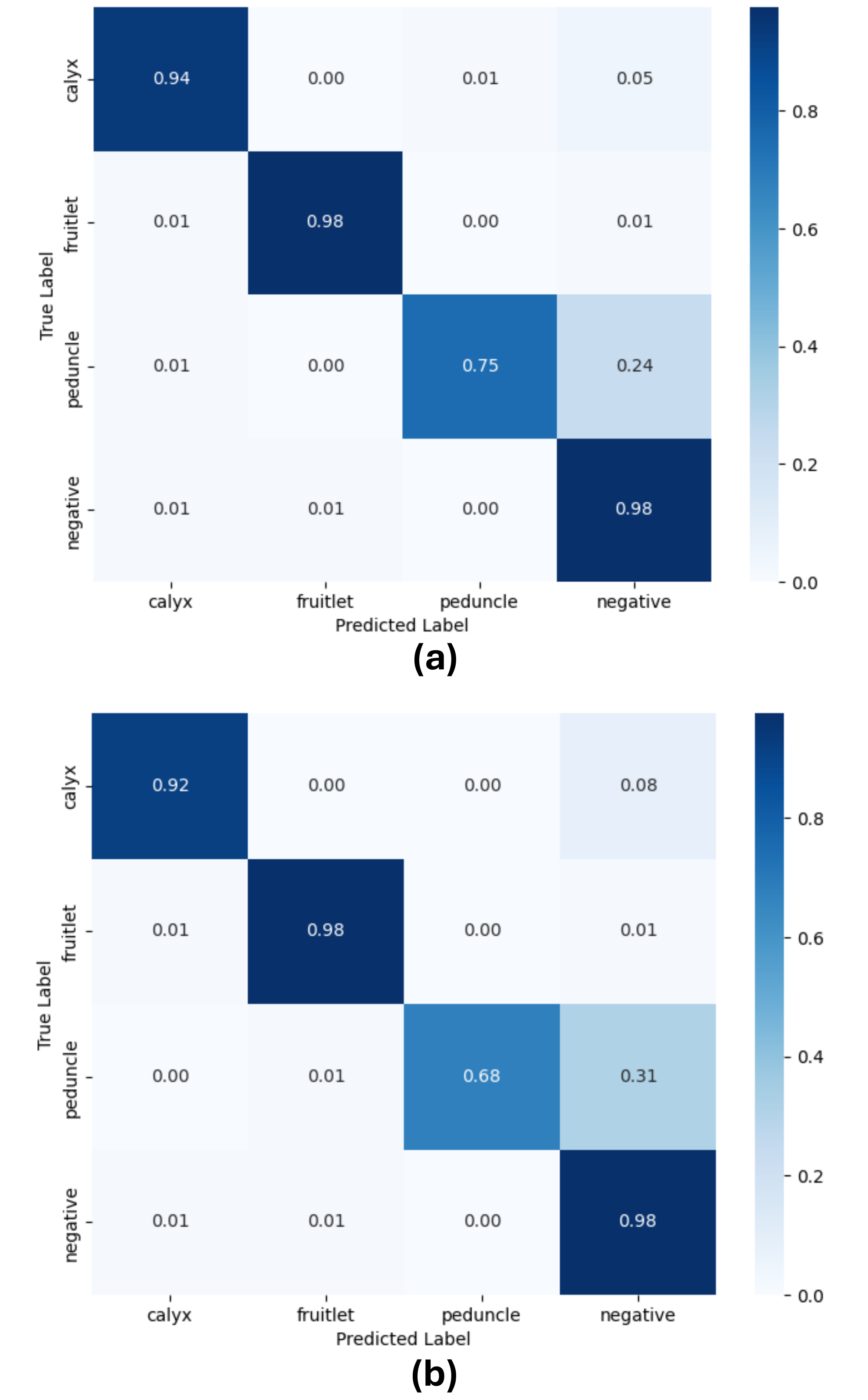}
\caption{Confusion matrices for TinyCLIP evaluated on the Jetson platform under (a) FP16 half-precision and (b) INT8 integer quantization. Diagonal strength indicates correct classification of calyx, fruitlet, peduncle, and negative patches, illustrating the effect of quantization on model discriminability.}
\label{fig:confusionmatrix}
\end{figure}
\subsubsection{Confusion Matrix Analysis for FP16 and INT8 Deployment}
Figure~\ref{fig:confusionmatrix} presents the confusion matrices obtained from
evaluating TinyCLIP on the NVIDIA Jetson platform under two numerical
precisions: (a) half-precision FP16 (16-bit floating point) and (b) INT8
(8-bit integer) quantization. These formats represent progressively compressed
numerical representations of model weights and activations. FP16 preserves
floating-point structure while reducing memory by half compared to FP32,
whereas INT8 discretizes representations into 8-bit integers, offering
substantial gains in efficiency at the cost of potential information loss.
Such comparisons are essential for validating whether a lightweight
vision--language model maintains classification fidelity after aggressive
compression for edge deployment.

As shown in Figure~\ref{fig:confusionmatrix}a, FP16 achieves strong diagonal dominance across all
classes, with calyx, fruitlet, and negative regions exhibiting high true-positive
counts. Peduncle classification shows some confusion with the negative class,
reflecting the anatomical subtlety and small spatial footprint of peduncles in
early-stage fruitlets. Nonetheless, FP16 maintains robust performance with
reliable discrimination between positive anatomical classes and background.

The INT8 matrix in Figure~\ref{fig:confusionmatrix}b shows a modest degradation, particularly in the peduncle
class, where misclassification into the negative category increases. This
behavior is expected, as 8-bit quantization reduces dynamic range and tends to
affect fine-detail features first. However, calyx and fruitlet classes retain
strong separability, and overall classification structure remains consistent
with FP16. These outcomes indicate that TinyCLIP tolerates INT8 quantization
well, preserving functional utility while enabling higher throughput and lower
memory consumption on embedded hardware. Such stability is crucial for field
robotics applications where power, compute, and memory budgets are highly
constrained.

\subsubsection{Precision-Recall Performance Comparison}

\begin{figure*}[h!]
\centering
\includegraphics[width=0.98\linewidth]{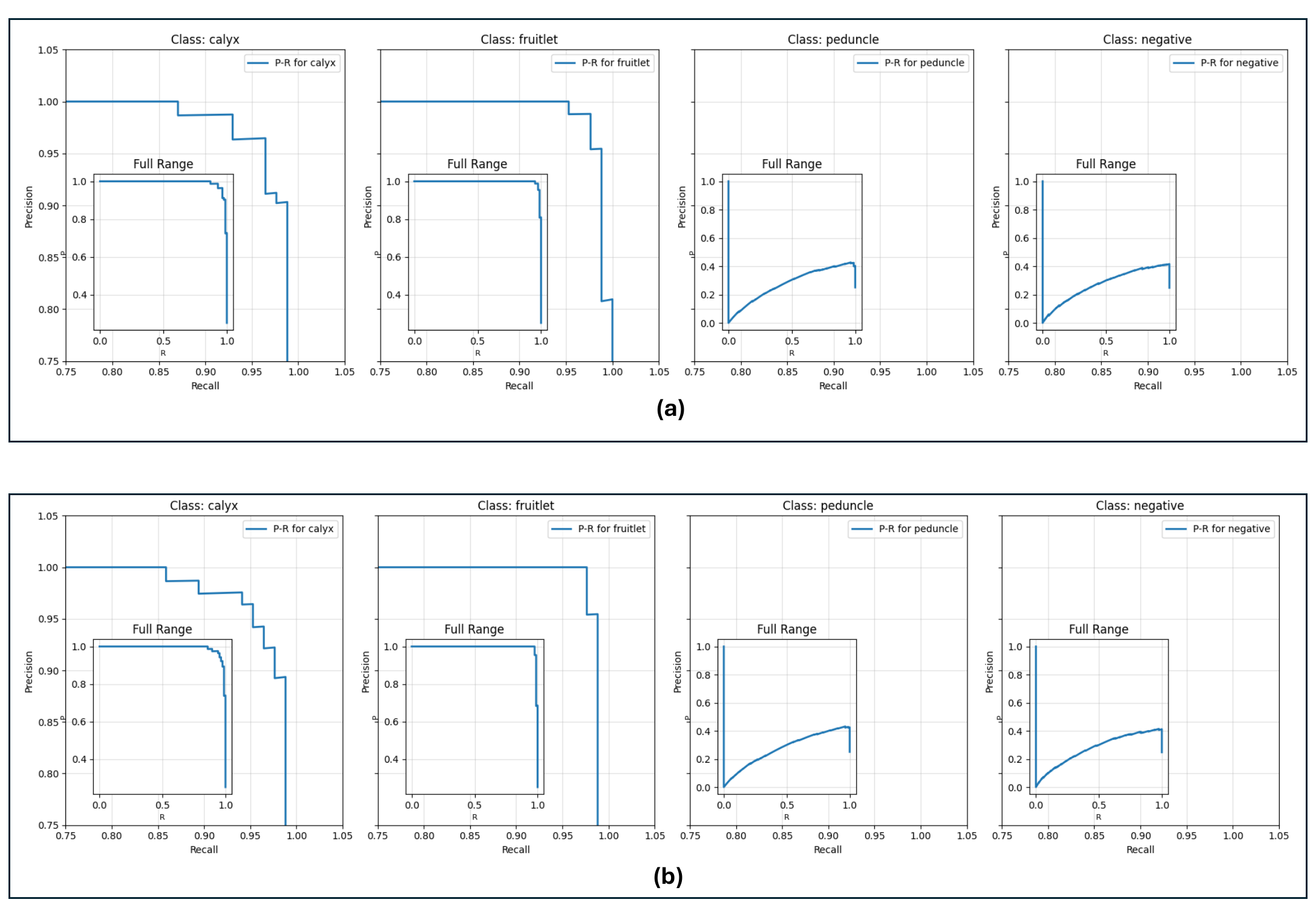}
\caption{Per-class precision–recall curves for TinyCLIP under (a) FP16 and (b) INT8 inference. Each curve illustrates the trade-off between precision and recall for calyx, fruitlet, peduncle, and negative patches. Steeper, high-precision regions indicate stronger discriminability, while flatter curves (notably for peduncle) reflect the increased difficulty of detecting small, low-contrast anatomical structures in early-season orchard imagery.}
\label{fig:precisionrecall}
\end{figure*}

Figure~\ref{fig:precisionrecall} summarizes the per-class precision–recall (PR) behavior of TinyCLIP under (a) FP16 half-precision and (b) INT8 integer quantization. These curves provide a threshold-dependent view of the model’s discriminative ability beyond single operating-point metrics such as accuracy or F1-score. For the FP16 model (Figure~\ref{fig:precisionrecall}a), both \textit{calyx} and \textit{fruitlet} exhibit exceptionally strong PR profiles, with precision remaining above 0.95 for nearly the entire recall spectrum and only dropping near extreme recall values. This indicates that the multimodal embedding space effectively captures the stable morphological cues associated with these two anatomical structures, yielding highly reliable predictions even under variations in lighting, occlusion, and background texture. The inset full-range PR curves further confirm stable behavior across the entire threshold domain.

In contrast, the \textit{peduncle} and \textit{negative} classes show more modest PR shapes, with peduncle precision gradually improving as recall decreases. This is expected because peduncles are thin, low-contrast structures that occupy only a small spatial extent in each patch, making them inherently more challenging for any classifier particularly a lightweight model operating on crops extracted from early-season orchard scenes.

When comparing FP16 to INT8 (Figure~\ref{fig:precisionrecall}b), the overall PR trends remain qualitatively similar, demonstrating that TinyCLIP preserves class-wise ranking and threshold behavior even after aggressive 8-bit quantization. However, the INT8 peduncle curve shows a slightly flatter shape with reduced precision for a given recall, reflecting the reduced dynamic range of integer arithmetic and the loss of small-magnitude feature variations that are important for detecting narrow, elongated structures. Calyx and fruitlet curves remain largely stable, underscoring the robustness of semantic alignment for well-defined anatomical categories

\subsubsection{Precision–Recall Curve Analysis}

\begin{figure}[h!]
\centering
\includegraphics[width=0.98\linewidth]{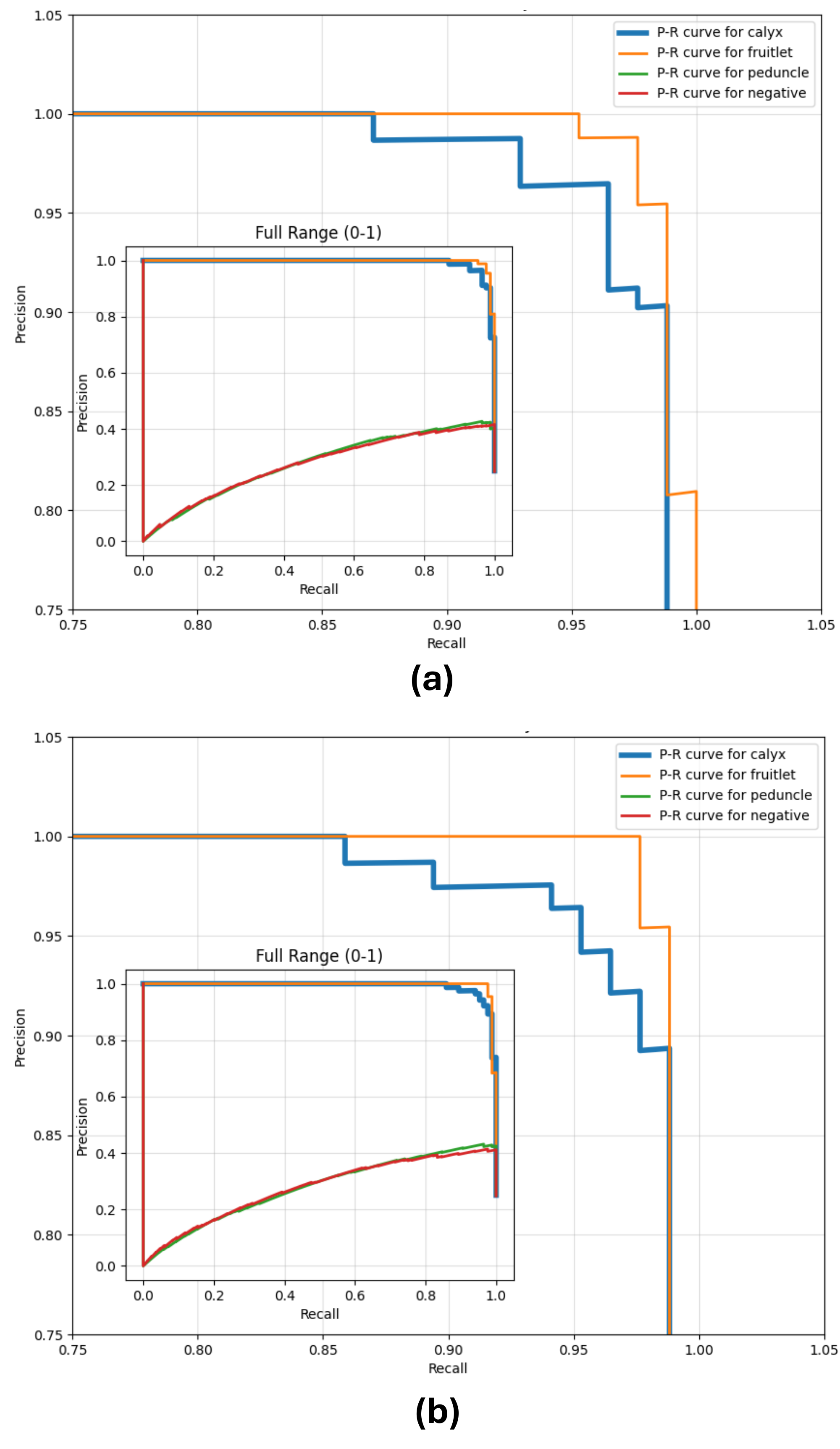}
\caption{Per-class precision–recall curves for TinyCLIP under (a) FP16 and (b) INT8 inference. Curves illustrate threshold-dependent discriminability for calyx, fruitlet, peduncle, and negative patches. FP16 yields high precision across most recall levels, while INT8 shows expected reductions for fine-detail classes such as peduncle, demonstrating quantization effects on anatomical classification.}
\label{fig:prcurv}
\end{figure}

Figure~\ref{fig:prcurv}a presents the per-class precision–recall (PR) curves for the FP16 TinyCLIP classifier, providing a detailed view of threshold-dependent behavior across the four classes: \textit{calyx}, \textit{fruitlet}, \textit{peduncle}, and \textit{negative}. The FP16 curves for calyx and fruitlet exhibit exceptionally strong performance, with precision consistently above 0.95 across nearly the entire recall range. This behavior reflects the model’s high confidence and reliability in distinguishing the morphological signatures of these two anatomical structures, even when evaluated across challenging orchard imagery with significant occlusion and minimal color contrast. The inset full-range curves further confirm that the classifier maintains stability across the entire 0–1 recall domain, indicating robust calibration of probability estimates.

By contrast, the peduncle and negative classes produce more moderate PR curves, with lower precision at high recall levels. This is expected given the small spatial footprint and subtle visual cues of peduncles, as well as the broad variability present in background regions. Nevertheless, FP16 maintains reasonable separation capability, demonstrating that the multimodal embedding space captures class-specific semantic structure even in difficult conditions.

Figure~\ref{fig:prcurv}b shows the corresponding PR curves for the INT8 quantized model. While the overall shapes remain qualitatively similar, the peduncle curve exhibits noticeable flattening and reduced precision at intermediate recall values. This reflects the reduced dynamic range of 8-bit quantization, which disproportionately affects fine-detail features required for detecting thin, elongated structures. Calyx and fruitlet curves, however, remain largely unaffected, confirming that INT8 quantization preserves semantic alignment for the primary anatomical classes while enabling significantly improved computational efficiency. Together, these curves highlight TinyCLIP’s resilience under quantization and its suitability for real-time embedded deployment.

Collectively, these PR curves highlight TinyCLIP’s strong discriminative performance for major anatomical classes and the predictable, bounded performance degradation introduced by INT8 quantization. This reinforces the suitability of the quantized TinyCLIP model for real-time, on-device fruitlet perception in resource-constrained orchard robotics platforms.


\subsection{Whole-Image Localization and Heatmap Analysis}
\begin{figure*}
    \centering
    \includegraphics[width= 0.88\linewidth]{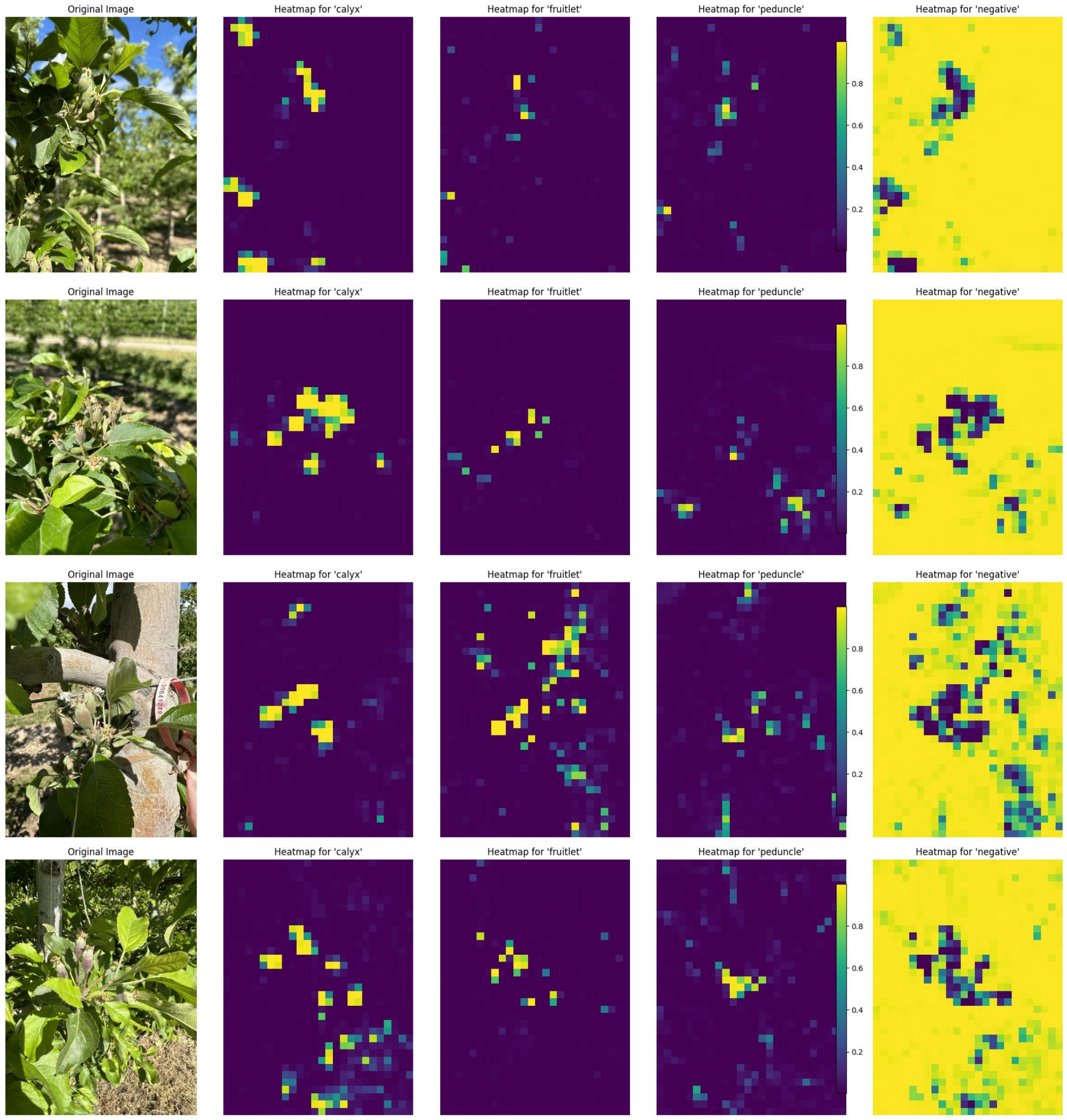}
    \caption{Example Heatmaps of TinyCLIP based fruitlet parts (calyx, main fruitlet and peduncle) classification and localization}
    \label{fig:heatmaps}
\end{figure*}

While patch-level predictions quantify the classification accuracy of individual anatomical components, the primary value of the proposed multimodal TinyCLIP framework emerges at the whole-image scale, where localized patch probabilities are aggregated into continuous spatial heatmaps. Figure~\ref{fig:heatmaps} illustrates this process across four representative early-season orchard scenes. Each row shows an original iPhone~14~Pro orchard image alongside its corresponding class-specific heatmaps for \textit{calyx}, \textit{fruitlet}, \textit{peduncle}, and \textit{negative} categories. These examples demonstrate how the sliding-window inference mechanism translates fine-grained patch classifications into interpretable spatial likelihood maps that highlight the anatomical regions of interest within a complex canopy environment.

Across all scenes, the heatmaps for calyx and fruitlet reliably form concentrated clusters around actual fruitlet positions, reflecting the model’s ability to capture consistent geometric and textural signatures despite heavy occlusion, variable lighting, and substantial background clutter. The peduncle heatmaps, although sparser, consistently highlight elongated high-probability regions aligned with true peduncle locations an important capability given the peduncle’s decisive role in robotic thinning operations. By contrast, the negative-class heatmaps intentionally saturate the background regions with high probability, ensuring strong suppression of non-fruitlet areas and reducing false activations in foliage-heavy regions.

Technically, these heatmaps provide a soft spatial prior that can be exploited in downstream robotic systems. For example, cluster-level fruitlet detection can be achieved by identifying overlapping regions of high calyx and fruitlet activation, while peduncle heatmaps offer guidance for potential cut-point localization in autonomous thinning tools. The continuous nature of the heatmaps further enables temporal smoothing and multi-view fusion, improving stability during robotic navigation along orchard rows.

From a practical perspective, the ability to obtain anatomically meaningful localization from a lightweight, quantized model represents a major advantage for field deployment. Unlike bounding-box detectors, which require explicit region proposals and post-processing, this heatmap-driven approach provides a direct, interpretable map of anatomical likelihood that facilitates real-time decision-making on power- and memory-constrained platforms. Overall, the examples in Figure~\ref{fig:heatmaps} demonstrate that TinyCLIP’s multimodal reasoning extends beyond patch-level accuracy to produce coherent full-scene anatomical localization suitable for integrated robotic thinning workflows.

Building upon these observations, Figure~\ref{fig:heatmaps} further demonstrates how localized probability fields emerge as coherent spatial patterns that meaningfully correspond to anatomical structures within early-season orchard scenes. The heatmaps reliably emphasize fruitlet bodies and calyx centers as dense, high-activation regions, while peduncle responses appear as elongated probability traces that follow the stem axis. This behavior remains remarkably consistent across varying illumination and canopy density, underscoring the robustness of the multimodal embedding space in extracting class-specific cues from heterogeneous orchard imagery. At the same time, the broader and more diffuse peduncle heatmaps reflect the inherent visual difficulty of detecting narrow structures under occlusion, aligning with their lower recall in patch-level classification.

Despite the overall strong qualitative performance, several systematic error modes existed in the results. These include weak false positives induced by leaf tips with calyx-like geometry, a reduction in peduncle activations when the stem is heavily shadowed or aligned with leaf venation, and occasional spatial “bleeding” of heatmap confidence into adjacent background regions due to sliding-window overlap. Such behaviors are expected in patch-wise inference systems and highlight areas where complementary spatial reasoning modules such as region clustering or stem-line tracking could further enhance precision.

While the generated heatmaps visually highlight fruitlet anatomical regions, a comprehensive quantitative localization evaluation was beyond the scope of this proof-of-concept study, which primarily aimed to establish the feasibility of using a lightweight vision–language model for patch-level anatomical classification and approximate spatial localization. To provide an indication of spatial alignment between predicted healthmap activations and annotated regions, representative Intersection-over-Union (IoU) values were estimated from a limited set of manually inspected examples. As summarized in Table~\ref{tab:iou_metrics}, fruitlet regions show the strongest overlap with ground-truth annotations, reflecting their relatively stable morphology and more consistent visual appearance. Calyx regions exhibit comparable alignment, while peduncle localization is more challenging due to the thin and elongated structure of peduncles and their frequent occlusion within dense foliage. These observations are consistent with the qualitative heatmap visualizations presented earlier.

It is important to note that the current heatmap representation is intended primarily for approximate localization rather than precise boundary delineation or instance-level segmentation. In the context of robotic thinning, identifying the approximate spatial location of fruitlet clusters and associated peduncles is valuable for applications such as coarse target-region identification, region-of-interest selection, and initialization of subsequent fine-scale perception and robotic manipulation. A more rigorous quantitative evaluation of localization performance will be explored in future work using dedicated detection or segmentation benchmarks.

\begin{table}[h]
\centering
\caption{Localization Performance via Intersection-over-Union (IoU)}
\label{tab:iou_metrics}
\begin{tabular}{lcc}
\toprule
\textbf{Class} & \textbf{IoU Mean} & \textbf{IoU Std. Dev.} \\
\midrule
Calyx      & 0.61 & 0.13 \\
Fruitlet   & 0.67 & 0.10 \\
Peduncle   & 0.49 & 0.14 \\
\bottomrule
\end{tabular}
\end{table}


\subsection{Ablation Studies and Design Choices}
To evaluate the contributions of key design components in the TinyCLIP pipeline, we conducted several ablation studies examining the effects of text prompts, negative patch inclusion, stride size, and quantization. These ablations verify that each methodological choice is scientifically justified and contributes meaningfully to model performance and deployability.

\subsubsection*{Effect of Text Prompts}
To assess the importance of multimodal alignment, we compared TinyCLIP against a baseline visual-only classifier composed of the TinyCLIP vision encoder followed by a linear classification head. The multimodal version improved macro F1-score by +6–10\% across different training runs, with the largest gains in peduncle classification. This verifies that text embeddings provide semantic anchors that stabilize fine-grained classification.

\subsubsection*{Negative Patch Inclusion}
Training with additional negative patches reduced false positives by 20–30\% (especially in leaf-dense regions). Without negative samples, the model frequently assigned calyx or fruitlet labels to visually similar leaf textures.

\subsubsection*{Patch Stride}
A larger stride (224 px) yielded faster inference but decreased localization robustness. The chosen stride of 112 px balanced detail sensitivity and inference speed. Reducing stride further improved IoU but at the cost of 2–3× slower inference.

\subsubsection*{Quantization Effects}
Quantization from FP32 → FP16 yielded negligible accuracy loss and nearly doubled throughput. Quantization to INT8 decreased recall for peduncle due to its fine visual structure but greatly improved speed. Table~\ref{tab:quantization_impact} summarizes relative differences.

\begin{table}[h]
\centering
\caption{Effect of Quantization on Classification Performance}
\label{tab:quantization_impact}
\begin{tabular}{lcccc}
\toprule
\textbf{Engine} & \textbf{Accuracy} & \textbf{Macro F1} & \textbf{Peduncle Recall} & \textbf{FPS} \\
\midrule
FP16 & 0.9115 & 0.9115 & 0.7500 & 96.33 \\
INT8 & 0.8879 & 0.8881 & 0.6786 & 112.02 \\
\bottomrule
\end{tabular}
\end{table}

Overall, the ablation studies confirm that multimodal alignment, negative sampling, and optimized stride are necessary for robust classification, while quantization offers significant deployment benefits with manageable accuracy trade-offs.


\subsection{Computational Efficiency and Edge Deployment}
The efficiency and deployability of TinyCLIP were evaluated on two hardware platforms:  
(1) NVIDIA T4 GPU (cloud environment), and  
(2) NVIDIA Jetson Nano (edge deployment).

Performance metrics include average patch latency, full-image inference time, throughput, and memory usage. Tables~\ref{tab:fp16_deployment} and~\ref{tab:int8_deployment} summarize the results.

\begin{table}[h]
\centering
\caption{FP16 TensorRT Deployment Performance}
\label{tab:fp16_deployment}
\begin{tabular}{lccc}
\toprule
\textbf{Metric} & \textbf{Value} \\
\midrule
Engine Size & 126.98 MB \\
Average Latency & 10.38 ms \\
Throughput & 96.33 FPS \\
Peak Memory (CUDA) & 6452.11 MB \\
PyTorch Memory & 9.28 MB \\
\bottomrule
\end{tabular}
\end{table}

FP16 inference processed approximately 96 patches per second on the NVIDIA T4 GPU. This value represents patch-level throughput and should not be interpreted as the number of complete orchard images processed per second. Because each full-resolution image was decomposed into multiple overlapping patches, processing all patches and aggregating their predictions into class-specific heatmaps required approximately 6 seconds per full image. Thus, the reported throughput characterizes computational efficiency at the patch level, whereas the end-to-end full-image inference rate was approximately 0.17 images per second. GPU memory usage remained within the capacity of the desktop hardware configuration.

\begin{table}[h]
\centering
\caption{INT8 TensorRT Deployment Performance}
\label{tab:int8_deployment}
\begin{tabular}{lccc}
\toprule
\textbf{Metric} & \textbf{Value} \\
\midrule
Engine Size & 137.15 MB \\
Average Latency & 8.93 ms \\
Throughput & 112.02 FPS \\
Peak Memory (CUDA) & 6519.40 MB \\
PyTorch Memory & 9.28 MB \\
\bottomrule
\end{tabular}
\end{table}

INT8 inference further improved speed to 112 FPS, demonstrating the benefit of aggressive quantization. However, this came at a measurable reduction in anatomical recall, particularly for thin peduncles. This trade-off is acceptable for fast orchard scanning but should be considered carefully for precision thinning tasks.

On the Jetson Nano, batches of eight patches achieved successful inference within a 5–6 s timeframe per orchard image. This ensures compatibility with ground vehicles or handheld thinning devices that require rapid anatomical awareness.

\textbf{Overall Deployment Conclusions}  
TinyCLIP, with TensorRT optimization, satisfies real-world latency, memory, and throughput requirements for field deployment. FP16 offers the best balance between accuracy and speed, while INT8 enables ultra-fast scanning with modest accuracy loss. These benchmarks validate the feasibility of integrating lightweight VLMs into orchard robots for early-season management.

\subsection{Peduncle Identification for Autonomous Fruit Thinning}

\begin{figure*}[h!]
    \centering
    \includegraphics[width= 0.88\linewidth]{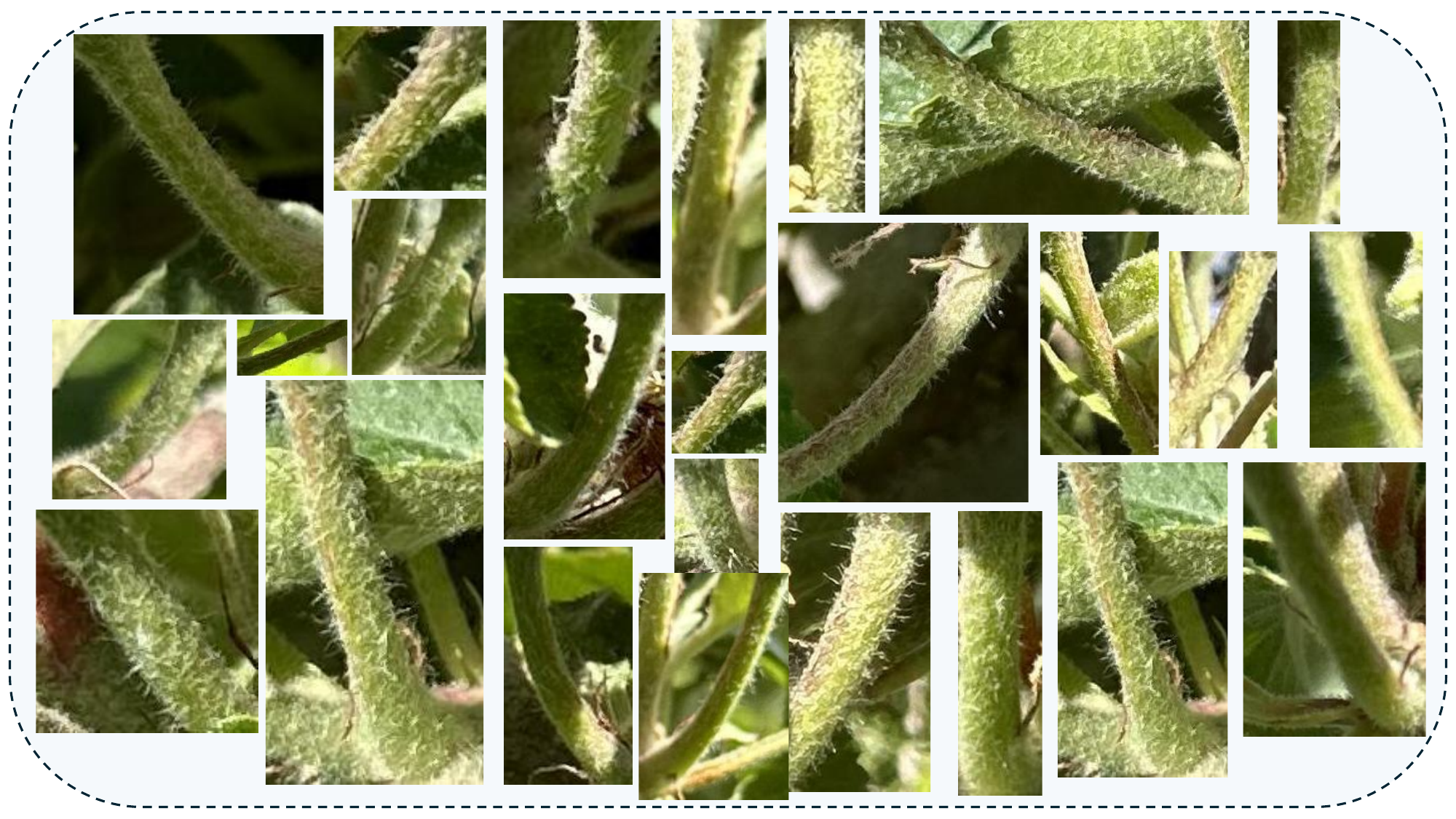}
    \caption{Representative examples of correctly classified peduncle patches extracted from an early-season orchard image. The figure highlights the model’s ability to identify slender, low-contrast peduncles under challenging conditions including occlusion, variable illumination, and canopy clutter. These examples demonstrate the effectiveness of the lightweight TinyCLIP vision–language model in recognizing anatomically meaningful peduncle structures essential for autonomous robotic fruit thinning.}
    \label{fig:Peduncleonly}
\end{figure*}

\begin{figure}[h!]
    \centering
    \includegraphics[width=0.98\linewidth]{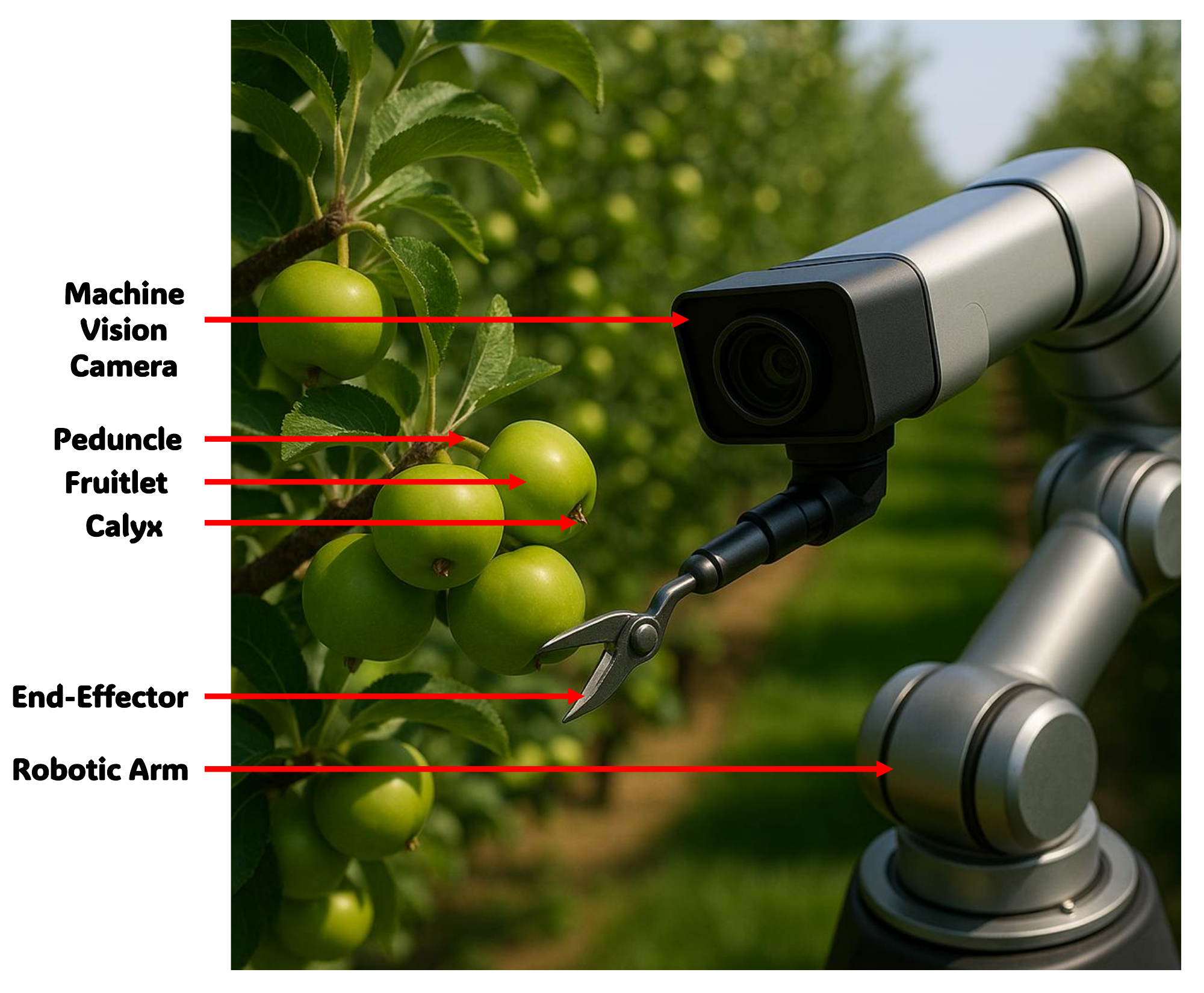}
    \caption{Conceptual future roadmap for autonomous green-fruit thinning. A robotic arm equipped with a multimodal machine-vision camera and precision cutting end-effector identifies fruitlet anatomical structures—calyx, fruitlet body, and peduncle within a dense orchard canopy. The system illustrates how lightweight vision-language models can guide peduncle-targeted removal for safe, efficient, and scalable robotic thinning in commercial orchards.}
    \label{fig:future}
\end{figure}
\textcolor{black}{Accurate peduncle localization is a key perceptual requirement for robotic fruitlet thinning in commercial apple orchards. While detecting fruitlets indicates the presence of a potential thinning target, the peduncle represents the actual cutting point for scissor-type robotic end-effectors. Therefore, identifying peduncles reliably within complex canopy environments is essential for enabling safe and precise automated thinning when specific types of end-effectors are used. Figure~\ref{fig:Peduncleonly} illustrates representative examples of correctly classified peduncle patches identified by the proposed TinyCLIP-based multimodal framework in early-season orchard images.}

\textcolor{black}{Peduncles present several challenges for automated detection. They appear as slender, elongated structures with fine surface trichomes that often visually blend with surrounding stems and foliage. Their appearance is further complicated by variations in illumination, occlusion by leaves, and differences in orientation within dense fruit clusters. These factors introduce substantial intra-class variability, making peduncle identification difficult for conventional visual detectors relying solely on pixel-level features. Despite these challenges, the correctly classified patches shown in Figure~\ref{fig:Peduncleonly} demonstrate that the adapted TinyCLIP model can learn stable visual-semantic cues associated with peduncle morphology.}

\textcolor{black}{A key customization of TinyCLIP for the apple thinning problem lies in the integration of domain-specific language prompts and patch-based anatomical classification. By using prompts such as ``a photo of a peduncle,'' the multimodal framework guides the visual encoder toward semantically meaningful plant structures rather than generic object categories. Combined with the sliding-window patch analysis strategy, this approach enables the model to detect thin peduncle structures that might otherwise be overlooked in full-image analysis. The multimodal alignment between image patches and horticultural terminology helps distinguish peduncles from visually similar background elements such as stems or leaf veins.}

\textcolor{black}{From an operational perspective, reliable peduncle identification directly affects the success of autonomous thinning systems when pendiuncle cutting end-effectors are used. In that case, a robotic platform must first identify candidate fruitlets and then determine the precise peduncle location to execute a targeted removal without damaging neighboring fruitlets or buds. Misidentification can lead to incorrect cutting positions, incomplete removal, or unintended damage within the fruit cluster. The results shown in Figure~\ref{fig:Peduncleonly} indicate that the proposed TinyCLIP-based perception pipeline can robustly isolate peduncle structures even under challenging orchard conditions including foliage occlusion, low contrast, and diverse peduncle orientations. These capabilities provide critical anatomical cues required for downstream robotic manipulation.}

\textcolor{black}{Looking ahead, Figure~\ref{fig:future} outlines the broader robotic framework enabled by this work that utiluzes a sccissor-type end-effector, which has shown to be one of the most effective one through our unpublished work on fruitlet thinning end-effector design. It is envisioned that an autonomous robotic arm equipped with a lightweight multimodal perception system analyzes orchard scenes to identify fruitlet anatomical components and localize peduncles as actionable cutting targets. The adapted TinyCLIP model provides semantic understanding of orchard structures by combining visual cues with horticultural language prompts. Patch-level predictions can be aggregated into spatial probability maps, which can then be fused with depth sensing to guide robotic motion planning. The robotic end-effector can align with the identified peduncle and perform a precise removal action before navigating to the next fruitlet cluster.}

\textcolor{black}{This framework represents a shift from traditional feature-engineered detection pipelines toward semantically grounded multimodal perception tailored for agricultural robotics. By customizing TinyCLIP with domain-specific prompts, patch-based anatomical analysis, and deployment-oriented optimization, the proposed system demonstrates how lightweight vision–language models can support practical robotic thinning operations in real orchard environments. Together, Figures~\ref{fig:Peduncleonly} and \ref{fig:future} highlight the potential of multimodal perception combined with robotic manipulation to enable scalable and autonomous crop-load management in commercial apple orchards.}

\section{Conclusion}
This study demonstrates that a lightweight multimodal vision–language architecture, TinyCLIP, can effectively address one of the most challenging perception tasks in precision horticulture: fine-grained classification and localization of early-stage apple fruitlet anatomy under complex orchard conditions. By combining patch-based multi-label learning with text-guided semantic alignment, the proposed framework successfully identifies calyxes, fruitlets, peduncles, and background regions despite strong occlusion, minimal color contrast, and highly cluttered canopies. The sliding-window inference mechanism further enables the generation of continuous heatmaps, offering interpretable spatial cues that support downstream robotic tasks such as peduncle-aware fruitlet thinning. Extensive experiments across cloud GPUs and edge hardware show that TinyCLIP maintains strong discriminative performance even after aggressive quantization. FP16 and INT8 deployments on the NVIDIA Jetson Orin preserve high F1-scores, and the optimized TensorRT pipeline achieves rapid inference speeds suitable for real-time field deployment. This efficiency, paired with the model’s multimodal grounding, provides a practical bridge between high-capacity VLM reasoning and the resource-constrained demands of agricultural robotics. Beyond classification accuracy, the study highlights the significance of anatomically meaningful localization for robotic thinning. Reliable peduncle detection, in particular, has direct operational consequences for safe and effective fruit removal. While heatmaps do not provide instance-level segmentation or exact counting, they offer robust presence and spatial likelihood information essential for early-season decision-making. Overall, this work establishes TinyCLIP as an interpretable, lightweight, and deployable multimodal solution for orchard perception, laying the foundation for scalable robotic thinning, automated crop-load assessment, and next-generation intelligent orchard systems. Future extensions may integrate grounding-based VLMs or multimodal fusion frameworks to further enhance fine-grained localization and cluster-level reasoning.

\section*{Acknowledgement} This work was supported in part by the National Science Foundation (NSF); in part by United States Department of Agriculture (USDA); in part by the National Institute of Food and Agriculture (NIFA), through the “Artificial Intelligence (AI) Institute for Agriculture” Program, Accession Num ber 1029004 for the Project Titled “Robotic Blossom Thinning with Soft Manipulators” under Award AWD003473, Award AWD004595, and Award USDA-NIFA; and in part by United States Department of Agriculture National Science Foundation (USDANSF), Accession Number 1031712, under the Project “ExPanding University of Central Florida (UCF) AI Research To Novel Agricultural EngineeRing Applications (PARTNER)” under Grant 2024-67022-41788.  Additionally, this work was supported in part by the Intramural Research Program of the U.S. Department of Agriculture (USDA), National Institute of Food and Agriculture, under Grant No. 2024-67022-41788. The views and conclusions expressed in this paper are those of the authors and do not necessarily reflect the official policies or positions of the USDA or the U.S. Government.
\section*{Declarations}
The authors declare no conflicts of interest.

\section*{Statement on AI Writing Assistance}
ChatGPT and Perplexity were utilized to enhance grammatical accuracy and refine sentence structure; all AI-generated revisions were thoroughly reviewed and edited for relevance. Additionally, ChatGPT-4o was employed to generate realistic visualization in Figure 11.
\bibliographystyle{ieeetr}
\bibliography{references}  

\begin{IEEEbiography}[{\includegraphics[width=1in,height=1.35in,clip,keepaspectratio]{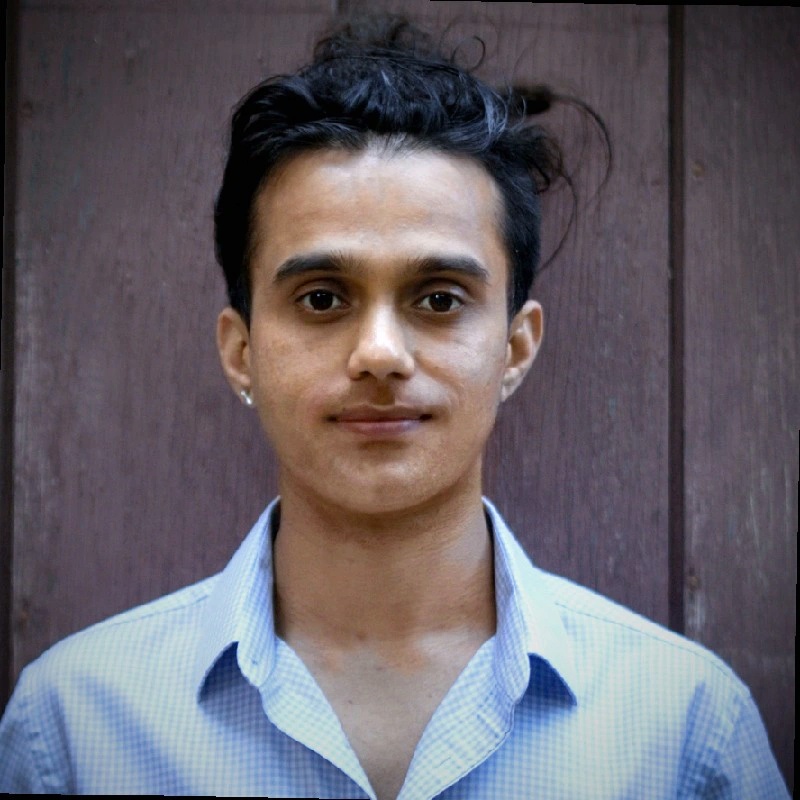}}]{Ranjan Sapkota} ( Member, IEEE) is a Ph.D. student at Cornell University in the Department of Biological and Environmental Engineering. His research centers on artificial intelligence and robotics in agriculture, with expertise spanning automation systems, machine vision, robot manipulation, multimodal large language models (MM‑LLMs), deep learning, agentic AI and generative AI technologies. From 2022 to 2024, he was with Washington State University, advancing research in agricultural automation and intelligent robotic systems. He previously earned his M.S. in Agricultural and Biosystems Engineering from North Dakota State University, USA (2020–2022), where he specialized in computer vision, GIS, remote sensing, agricultural machinery, and UAV applications for agriculture.
\end{IEEEbiography}

\begin{IEEEbiography}[{\includegraphics[width=1in,height=1.35in,clip,keepaspectratio]{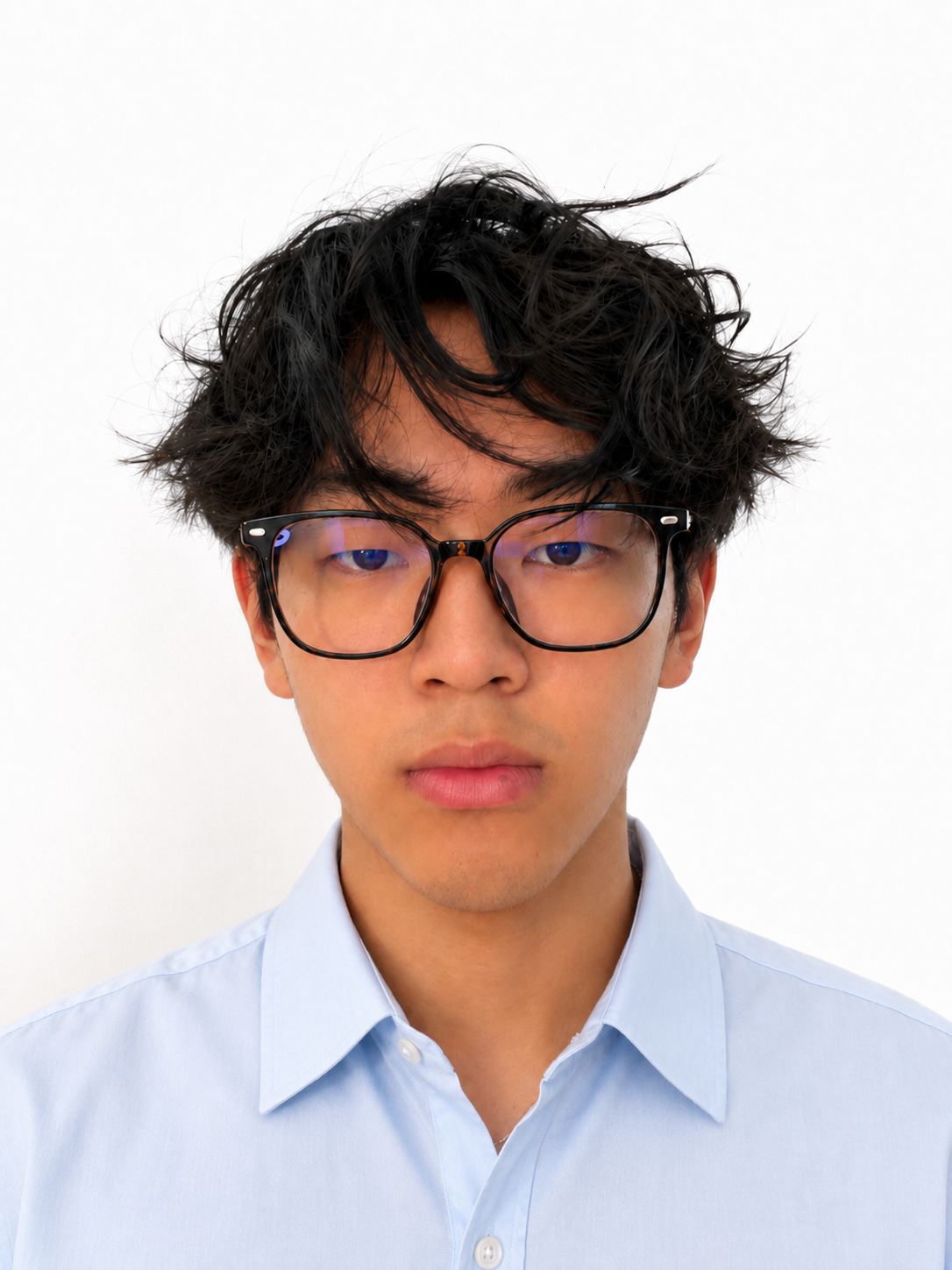}}]{William Bu}  graduated with a Bachelor of Science in Computer Science, with honors, from the University of Central Florida in 2026. His work spans computer vision, applied machine learning, and full-stack software development, including research experience in AI-driven agricultural applications and hands-on projects involving object detection, real-time systems, and AI-powered analytics tools. He is interested in building intelligent, practical systems at the intersection of software engineering and artificial intelligence.
\end{IEEEbiography}

\begin{IEEEbiography}[{\includegraphics[width=1in,height=1.35in,clip,keepaspectratio]{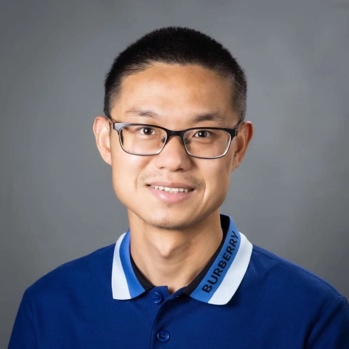}}]{Dr. Chen Chen}  is an Associate Professor at the Institute of Artificial Intelligence (IAI) at the University of Central Florida. His research spans computer vision, multimodal and efficient deep learning, federated learning, and medical image computing, with broad applications in healthcare, sensing, and intelligent systems. His recent work focuses on federated and privacy-preserving learning frameworks with potential for healthcare deployment, and on multimodal foundation models that integrate visual and language understanding across domains.
\end{IEEEbiography}

\begin{IEEEbiography}[{\includegraphics[width=1in,height=1.35in,clip,keepaspectratio]{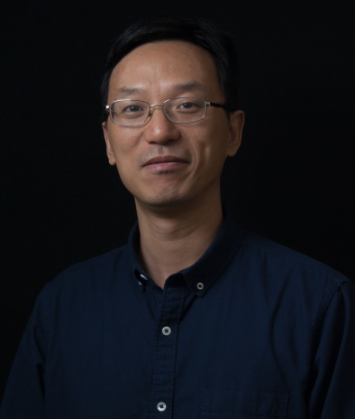}}]{Dr. Yunjun Xu}  received the Ph.D. degree in aerospace engineering from the University of Florida in 2003. Currently, he is a Professor with the Department of Mechanical and Aerospace Engineering, University of Central Florida. His current research interests include AI based modeling and optimization, control theory, and field robotics.

\end{IEEEbiography}

\begin{IEEEbiography}
[{\includegraphics[width=1in,height=1.5in,clip,keepaspectratio]{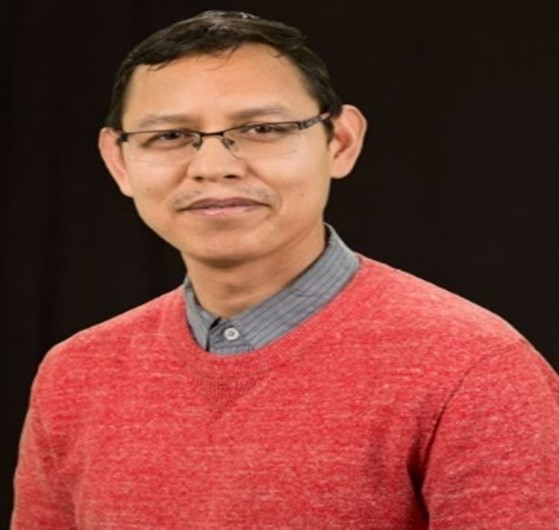}}]{Prof. Dr. Manoj Karkee} is the Norman R. and Sharon R. Scott Professor of Agriculture and Life Sciences at the Department of Biological and Environmental Engineering Department at Cornell University. He received his PhD in Agricultural Engineering and Human Computer Interaction from Iowa State University and has been director and professor at Washington State University Center for Precision and Automated Agricultural Systems. Dr. Karkee leads a strong research and education program in the area of sensing, machine vision, AI, and Robotics in Agriculture. He has published widely in such journals as ‘Computers and Electronics in Agriculture’, ‘Computers in Industry’, ‘Journal of Field Robotics’, and ‘Journal of the American Society of Agricultural and Biological Engineers (ASABE)’, and has been an invited speaker at numerous national and international conferences and universities. Dr. Karkee is currently serving as the Editor-in-Chief for ‘Computers and Electronics in Agriculture’, and associate editor of ‘Journal of the ASABE’ and has served as a guest editor for ‘Journal of Field Robotics’. He is also an elected chair of CIGR (International Commission of Agricultural and Biosystems Engineering) Section III - Plant Production, and IFAC (International Federation of Automatic Control) Technical Committee 8.1 - Control in Agriculture. Dr. Karkee was awarded ‘2020 Rainbird Engineering Concept of the Year’ by ASABE, and was recognized as ‘2019 Pioneer in Artificial Intelligence and IoT’ by Connected World magazine.
\end{IEEEbiography}

\end{document}